\documentclass[pdflatex,sn-mathphys-num]{sn-jnl}
\usepackage{graphicx}%
\usepackage{multirow}%
\usepackage{amsmath,amssymb,amsfonts}%
\usepackage{amsthm}%
\usepackage{mathrsfs}%
\usepackage[title]{appendix}%
\usepackage{xcolor}%
\usepackage{textcomp}%
\usepackage{manyfoot}%
\usepackage{booktabs}%
\usepackage{algorithm}%
\usepackage{algorithmicx}%
\usepackage{algpseudocode}%
\usepackage{listings}%
\usepackage{url}
\usepackage{comment}

\theoremstyle{thmstyleone}%
\theoremstyle{thmstyletwo}%

\theoremstyle{thmstylethree}%

\begin{document}

\title[Playing social deduction games with reinforcement fine-tuned large language models]{Playing social deduction games with reinforcement fine-tuned large language models}


\author[1]{\fnm{Lingzhe} \sur{Zhang}}\email{zhang.lingzhe@stu.pku.edu.cn}

\author[2]{\fnm{Yunpeng} \sur{Zhai}}\email{zhaiyunpeng.zyp@alibaba-inc.com}

\author*[1]{\fnm{Tong} \sur{Jia}}\email{jia.tong@pku.edu.cn}

\author[3]{\fnm{Kening} \sur{Zheng}}\email{kzhen28@uic.edu}

\author[1]{\fnm{Chiming} \sur{Duan}}\email{duanchiming@stu.pku.edu.cn}

\author[1]{\fnm{Minghua} \sur{He}}\email{hemh2120@stu.pku.edu.cn}

\author[2]{\fnm{Zhaoyang} \sur{Liu}}\email{jingmu.lzy@alibaba-inc.com}

\author[2]{\fnm{Bolin} \sur{Ding}}\email{bolin.ding@alibaba-inc.com}

\author[3]{\fnm{Philip S.} \sur{Yu}}\email{psyu@uic.edu}

\author*[1]{\fnm{Ying} \sur{Li}}\email{li.ying@pku.edu.cn}

\affil*[1]{\orgname{Peking University}, \orgaddress{\city{Beijing}, \country{China}}}

\affil[2]{\orgname{Alibaba Group}, \orgaddress{\country{China}}}

\affil[3]{\orgname{University of Illinois Chicago}, \orgaddress{\city{Chicago}, \state{Illinois}, \country{United States}}}


\abstract{Reinforcement fine-tuning (RFT) is increasingly used in applications where large language models (LLMs) interact with humans and other agents. Here we use social deduction games to study how RFT changes LLMs' social behaviour. We let fine-tuned and base LLM agents play hidden-role games that require hidden-state inference, social reading and vote steering. Our results show that LLM agents do not reliably acquire social-deduction ability by directly optimizing terminal win--loss outcomes, suggesting that final game results provide a sparse and noisy signal for socially interactive learning. However, RFT is particularly effective at improving social reading, including tasks that require agents to infer hidden roles from public discussion, update beliefs over time and predict other agents' future decisions. We further show that RFT can also improve social influence, including tasks that require agents to steer votes, team approvals and collective decisions, although these gains depend more strongly on behaviourally specific rewards and structured interaction settings. Finally, we show that LLMs' ability to play social deduction games can be further improved through multi-agent social-cognitive reinforcement fine-tuning, which combines social-reading and social-influence signals during same-side multi-agent training. These learned behaviours also receive more favourable human evaluations of strategic competence, persuasiveness and social usefulness. Together, these results enrich our understanding of how RFT changes LLMs' social behaviour and provide a step toward a behavioural learning theory for machine social intelligence.}


\keywords{Social Deduction Game, Reinforcement Learning, Large Language Model}

\maketitle

\section{Introduction}\label{sec:intro}

Large language models (LLMs) have achieved remarkable progress in recent years, demonstrating strong capabilities in language understanding~\cite{kuang2025natural}, reasoning~\cite{guo2025deepseek,zhang2026agentic,zhang2026hypothesize}, coding~\cite{el2025competitive,nam2024using} and decision-making across diverse domains~\cite{sarukkai2026self,zhang2025survey2,zhang2025microremed,liu2025ora}. These capabilities are increasingly turning LLMs from passive text generators into agents that act in interactive environments, including tool-use workflows~\cite{duan2026aims}, collaborative problem solving~\cite{zhang2025agentfm,hong2025cslparser,zhang2025adaptive,zhang2026towardsrootcause}, multi-agent simulations~\cite{tang2025synthesizing,wang2025user}, human--AI decision-making environments~\cite{gonzalez2026toward} and artificial societies~\cite{liu2026lmagent,tang2025gensim,ji2026leveraging}. At the same time, reinforcement fine-tuning (RFT) has become a central post-training mechanism for adapting pretrained LLMs for goal-directed agency~\cite{zhang2025survey,zheng2025learning}. By optimizing model behaviour with task-specific reward signals, RFT can improve reasoning quality~\cite{tan2026reason}, response behaviour~\cite{li2026think} and long-horizon task performance~\cite{zhai2025agentevolver,zhang2025thinkfl,zhang2026e2e,zhang2026feedback}. Together, these trends raise a central question: how does RFT shape the social reasoning of LLM agents when they interact with other agents?

Understanding how people learn to navigate social interactions, for example, how they infer others' intentions, build trust, adapt communication and influence group decisions, is a central topic across social cognition, social learning and behavioural game theory. These perspectives emphasize complementary aspects of social behaviour. Social cognition and theory of mind study how people infer others' beliefs, intentions and likely actions from limited behavioural evidence~\cite{call2008does}. Social learning theory examines how behaviour changes through observation, feedback and the consequences of others' actions~\cite{akers2015social}. Behavioural game theory further studies how people cooperate, coordinate, compete and adapt in strategic interactions~\cite{camerer2018behavioural}. Together, these perspectives provide a behavioural framework for studying how LLM-based agents acquire social competence through interaction: they must read other agents from observable behaviour, update their strategies from feedback and act in ways that shape collective outcomes.

\begin{figure*}[htbp]
	\centering
	\includegraphics[width=1\linewidth]{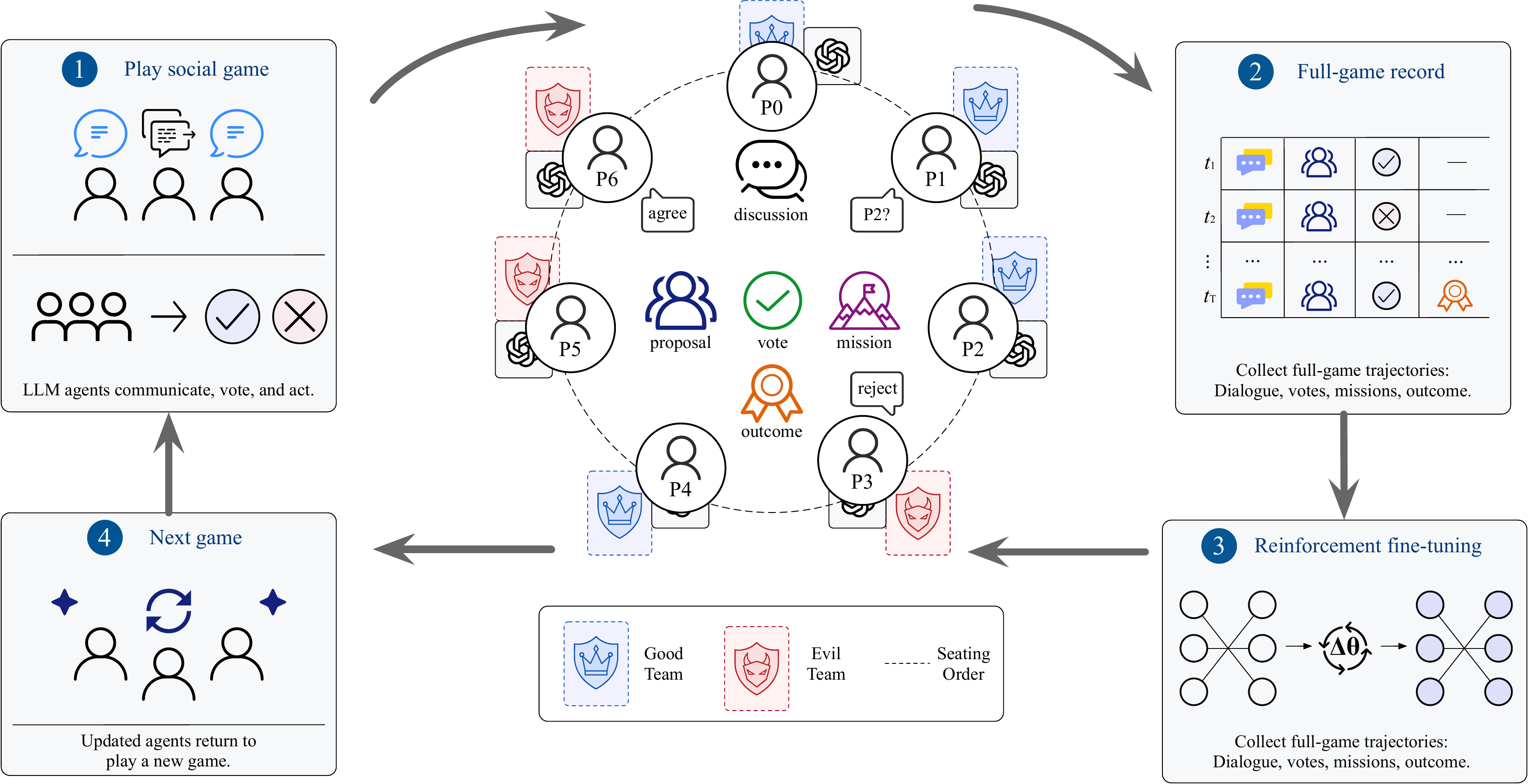}
	\caption{\textbf{Playing social deduction games with reinforcement fine-tuning in an example game of Avalon.} In step 1, LLM-controlled agents with hidden good or evil roles interact in an Avalon-like social deduction game through public discussion, team proposals, voting, missions and final outcomes. In step 2, the full-game trajectory, including dialogue, votes, mission results and outcome, is collected as a complete episode record. In step 3, reinforcement fine-tuning updates the agents from this full-game experience. In step 4, the updated agents return to play the next game, forming a closed learning loop across repeated social deduction games.}
	\label{fig:intro-example}
\end{figure*}

In this Article, we analyse how reinforcement fine-tuning changes LLM agents' social behaviour by letting fine-tuned agents play social deduction games against pretrained LLM agents. Social deduction games have been designed to study how agents reason and act when social identities are hidden, information is asymmetric, communication is public and decisions are made collectively. We focus on role-based hidden-information games in which players receive private identities, interact through public discussion, make proposals or votes, observe partial outcome signals and pursue team-level objectives (see Fig.~\ref{fig:intro-example} for an overview).

Analysing agents' performance across these games, we find that RFT improves social reasoning in a reward-dependent way. Fine-tuned agents perform particularly well on social reading tasks, including hidden-role inference from public discussion and prediction of other agents' future decisions. Direct outcome-only training, however, is unstable because terminal wins provide a sparse and entangled signal for dialogue, belief formation and group decisions. Social influence is not absent from learning: in controlled vote-steering and team-approval tasks, agents learn to move other players' decisions when the reward explicitly captures the intended downstream social effect. This contrast motivates our main methodological claim: social deduction should be decomposed into behaviourally grounded social-cognitive targets, and then recombined through a training protocol that respects the multi-agent structure of the game. We therefore study same-side multi-agent RFT, where all roles on a target side are trained from shared game rollouts and terminal win rate is used as an evaluation diagnostic rather than as the primary online reward. This design tests whether learned social-cognitive behaviours transfer from local diagnostics to full-game outcomes in LLM--LLM interaction.

Analysing agents' performance across these games, we find that RFT changes LLM agents' social behaviour in a structured but reward-dependent way. Directly optimizing terminal win--loss outcomes does not reliably produce general social-deduction ability, because final game results entangle hidden-role inference, public communication, vote coordination and other agents' responses into a sparse feedback signal. By contrast, when the reward targets specific social-cognitive functions, RFT produces clearer improvements. Fine-tuned agents become better at social reading, including inferring hidden roles from public discussion, updating beliefs as new evidence accumulates and predicting other agents' future decisions. They can also acquire social influence: in vote-steering and team-approval tasks, trained agents learn to move other players' decisions toward strategically relevant targets, although these effects are more sensitive to the reward definition and interaction setting than social-reading gains. We then ask whether these component improvements can be converted into stronger full-game play through multi-agent social-cognitive reinforcement fine-tuning, which trains agents with social-reading and social-influence signals in complete game trajectories. Finally, we evaluate whether the resulting behaviours are perceived by human observers as more strategic, persuasive and socially useful.

\section{Results}\label{sec:result}

Using Qwen2.5-7B-Instruct, Qwen2.5-14B-Instruct, Qwen3-4B-Instruct, GLM-4-9B-Chat, and Llama-3.1-8B-Instruct, we study whether reinforcement fine-tuning can improve LLM agents' social capabilities in social deduction games, including One Night Werewolf, Werewolf/Mafia, Avalon, Secret Hitler, and Spyfall. Our main controlled experiments focus on Avalon and One Night Werewolf, where we examine two focal capabilities: social reading and social influence.

\subsection{Analysing RFT across social deduction games}

We start our experiments by evaluating reinforcement fine-tuning across major families of social deduction games. We group these games into three macro-families according to how hidden information becomes public evidence: Elimination, Institution, and Asymmetry. Elimination games reveal hidden roles through discussion, accusations, voting, kills, and removals; Institution games reveal hidden allegiances through structured procedures such as team proposals, missions, elections, and policies; and Asymmetry games reveal hidden private knowledge or culpability through questions, answers, clues, and consistency checks. For each macro-family, we instantiate two representative games: Werewolf/Mafia and One Night Werewolf for Elimination, Avalon and Secret Hitler for Institution, and Spyfall and Deception: Murder in Hong Kong for Asymmetry. We show example game traces and decision points for each macro-family in Fig.~2.

\begin{figure}[h]
	\centering
	\includegraphics[width=\linewidth]{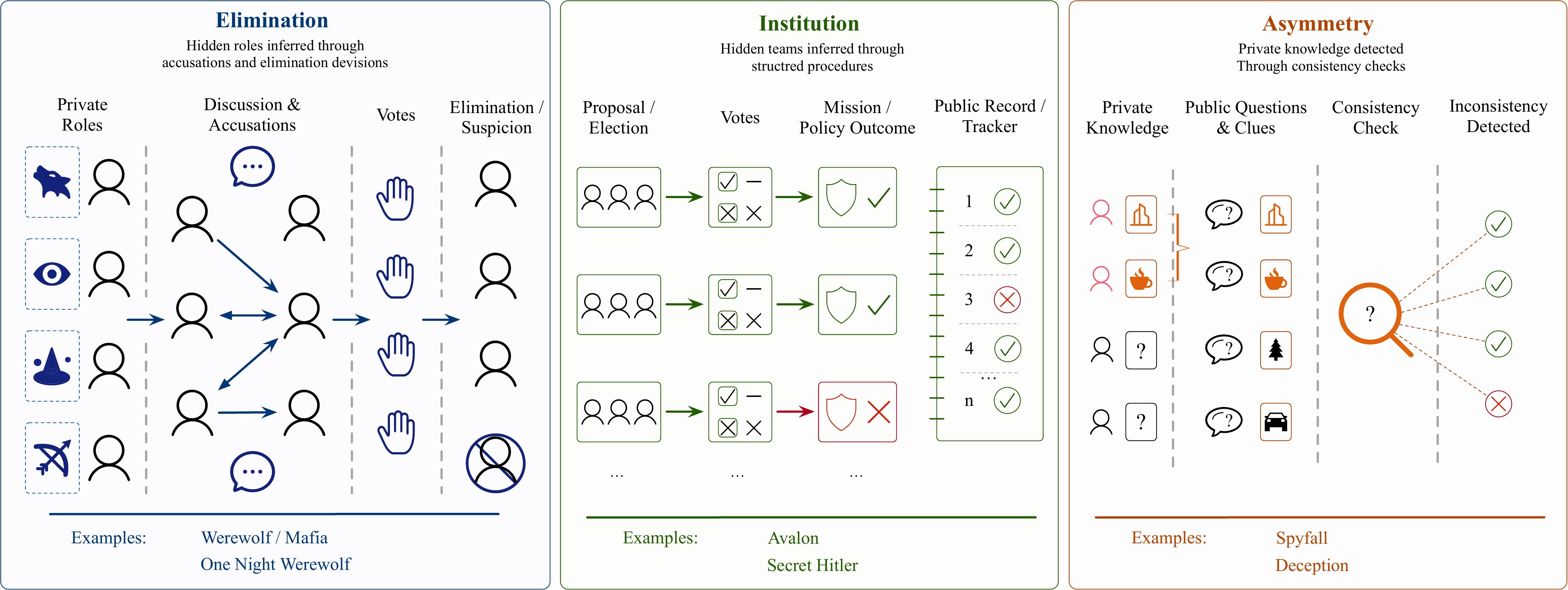}
	\caption{
		\textbf{Macro-families of social deduction games.}
	}
	\label{fig:social-deduction-game}
\end{figure}

For each game, we construct a compact training environment in which the fine-tuned model controls one focal player, while the other players are implemented as fixed pretrained LLM agents and are not updated during RFT. During training, the focal player is sampled across available roles and teams whenever the game permits, rather than being tied to a single role. This mixed-role setup prevents the policy from overfitting to a fixed perspective and requires it to act under different hidden states, including innocent, adversarial, and special-information roles. Each rollout consists of a complete game episode, including private role assignment, public discussion or institutional actions, and the final decision stage such as a vote, mission outcome, policy outcome, or accusation.

\begin{figure}[h]
	\centering
	\includegraphics[width=\linewidth]{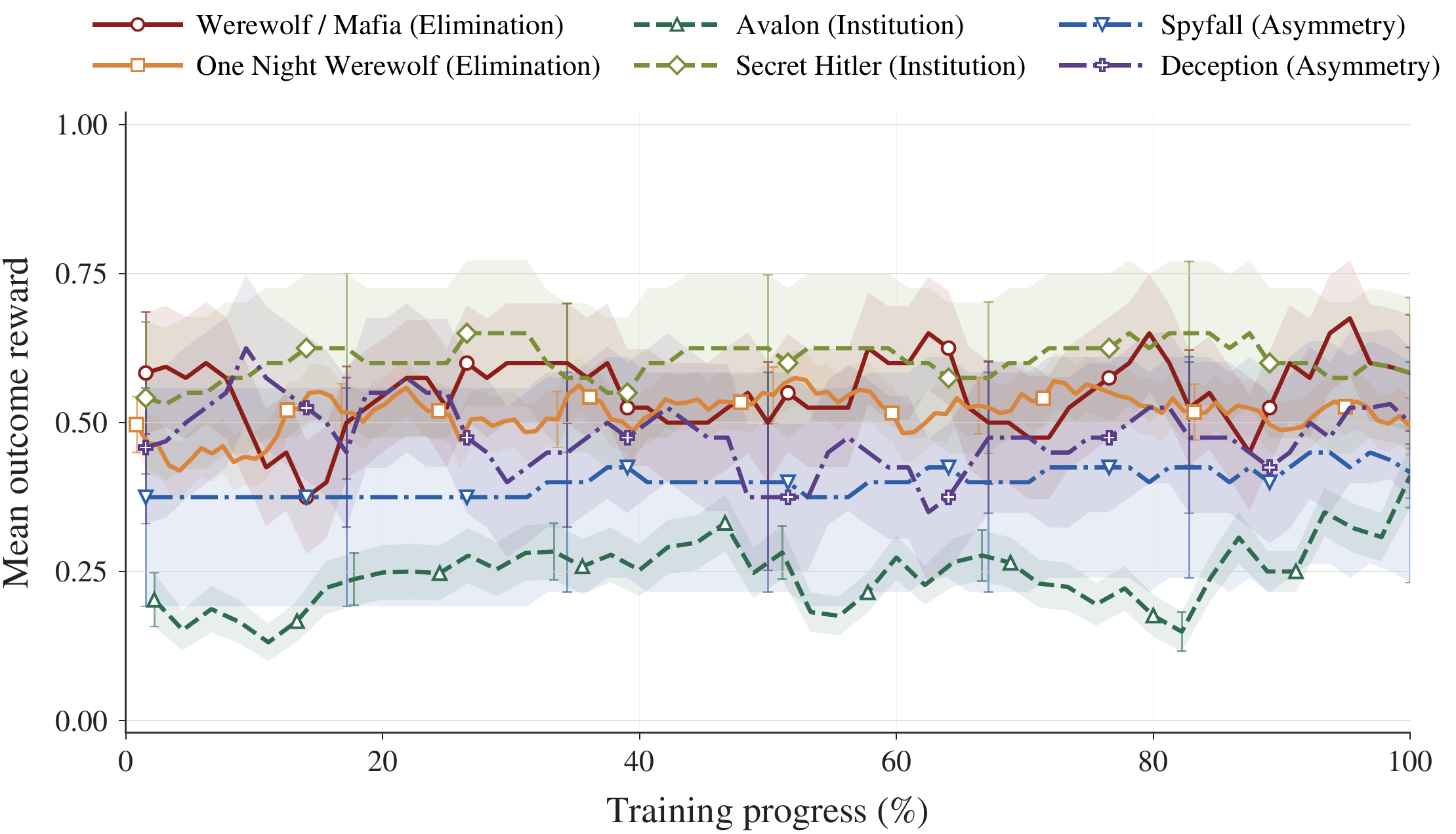}
	\caption{
		\textbf{RFT training dynamics across social deduction game families.} We report the mean episode-level outcome reward during training for representative games from the Elimination, Institution, and Asymmetry families. The reward is defined by whether the focal agent's side achieves its game-specific objective at the end of the episode. Shaded bands and capped error bars indicate the standard error across rollouts within each training step.
	}
	\label{fig:social-deduction-reward-curves}
\end{figure}

We use outcome-based rewards for these cross-game experiments. The reward is $1$ if the focal player achieves its side's objective at the end of the game and $0$ otherwise. Thus, innocent or investigator-side players are rewarded for correctly identifying hidden adversaries or helping their side win, while adversarial players are rewarded for surviving, misleading the group, or causing the wrong collective decision, depending on the game. We then apply RFT to the focal policy using these episode-level rewards and report the mean reward over training in Fig.~\ref{fig:social-deduction-reward-curves}.

Across games, the reward curves show that direct RFT on social deduction outcomes is challenging. Rather than producing uniformly monotonic improvements, training often leads to high-variance and game-dependent dynamics. This is expected because success in social deduction depends on long-horizon credit assignment over dialogue, hidden-role inference, and multi-agent strategic responses, while the final reward is sparse and binary.

The Institution games provide the clearest positive signal. Secret Hitler achieves the strongest overall reward among the completed runs, with a final mean reward of approximately $0.63$ and an average reward of approximately
$0.61$ across training. This suggests that RFT can exploit structured institutional feedback when actions such as elections, policies, or votes create relatively clear evidence about hidden teams. Avalon, however, remains substantially harder: its reward starts around $0.23$ and ends around $0.39$ over the available training steps, with an average reward of only about $0.25$. This contrast indicates that even within the same macro-family, role-sensitive reasoning and asymmetric strategic objectives can make outcome-only RFT much less stable.

The Elimination games show more volatile dynamics. Werewolf/Mafia reaches a high peak reward of approximately $0.88$, but the reward does not remain stable and ends around $0.38$. This suggests that RFT can discover locally useful accusation or voting behaviours, but these behaviours do not consistently translate into robust social deduction performance. One Night Werewolf is more stable, with an average reward of approximately $0.52$ and a
final reward close to its initial value. However, its curve is also non-monotonic and does not show sustained improvement, suggesting that even in a shorter elimination game with fewer interaction rounds, outcome-only RFT does not reliably improve social deduction performance.

The Asymmetry games also show limited direct improvement. Spyfall remains relatively flat, with an average reward of approximately $0.40$ and a final reward around $0.50$, indicating only modest learning from question-answer based private-information detection. Deception exhibits similarly noisy training dynamics: although it reaches a peak reward of approximately $0.75$, its final reward returns to about $0.50$, with an average reward of approximately $0.48$ across training. This suggests that RFT can occasionally discover useful clue-based accusation strategies, but these gains are not sustained throughout training. These games require the agent to connect sparse clues and linguistic consistency checks to hidden culpability, making the reward signal especially indirect.

Overall, Fig.~\ref{fig:social-deduction-reward-curves} suggests that outcome-based RFT alone does not reliably produce broad improvements in social deduction ability. The strongest gains appear in games with more structured procedural feedback, while elimination- and clue-centered games exhibit higher variance and weaker final improvements. These results motivate our subsequent controlled analyses in Avalon and One Night Werewolf, where we isolate more specific social capabilities, such as social reading and social influence, rather than relying only on aggregate game outcomes.

\subsection{Social reading and influence}

The cross-game results above show that outcome-based RFT does not uniformly improve overall social deduction performance. We therefore turn to controlled experiments that isolate two core social capabilities required by these games: \emph{social reading}, the ability to infer hidden social states from others' behaviour, and \emph{social influence}, the ability to shape other players' beliefs and decisions through communication. This allows us to examine when RFT fails on aggregate game outcomes, and when it can still improve specific social skills.

\subsubsection{Social Reading}

We study social reading along three increasingly game-integrated settings. First, we test whether agents can infer hidden states such as roles or teams from observed dialogue and actions. Second, we examine whether agents can update these beliefs as new evidence accumulates over the course of a game. Third, we analyse a role-specialized case in Avalon, where the Assassin's success depends strongly on identifying Merlin. Together, these experiments separate static inference, sequential belief updating, and the use of reading for role-specific game success.

\paragraph{Hidden-state inference.}

We first isolate social reading as a static hidden-state inference problem. In this setting, the agent observes a partial game transcript and is asked to predict latent social variables that are not directly visible from the dialogue alone. For Avalon, the target is each player's hidden team alignment. For Werewolf/Mafia, the target is the hidden wolf set. This evaluation removes the need to optimize full-game strategy and instead asks whether RFT can improve the agent's ability to infer social state from public behaviour, claims, votes, and outcomes.

\begin{figure}[h]
	\centering
	\includegraphics[width=\linewidth]{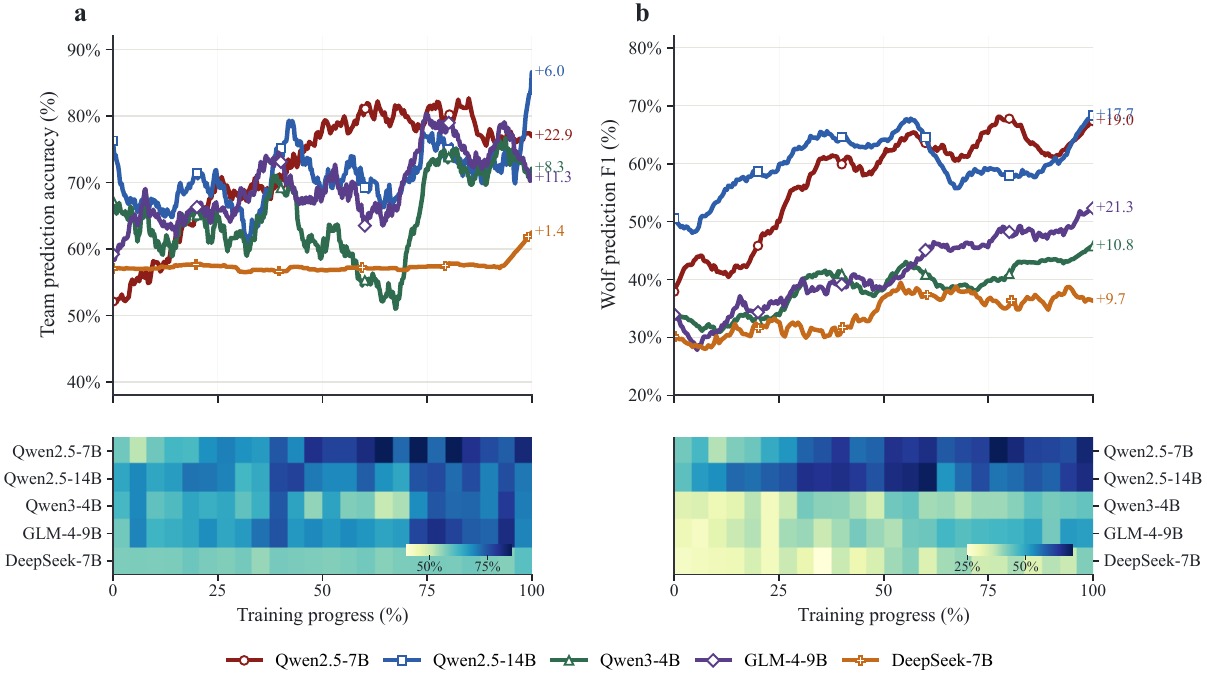}
	\caption{
		\textbf{RFT improves hidden-state inference across social deduction games.}
		We train agents on controlled social-reading tasks and report hidden-state prediction accuracy over training. 
		\textbf{a,} In Avalon, agents infer each player's hidden team alignment from public game transcripts. 
		\textbf{b,} In Werewolf/Mafia, agents infer the hidden wolf set from dialogue and voting evidence. 
		Across completed Qwen, GLM, and DeepSeek runs, RFT improves hidden-state inference, indicating that reinforcement fine-tuning can strengthen social reading even when aggregate outcome-based game training is noisy.
	}
	\label{fig:social-reading-hidden-state-inference}
\end{figure}

Fig.~\ref{fig:social-reading-hidden-state-inference} shows that RFT improves hidden-state inference in both games across multiple model families. In Avalon, Qwen2.5-7B improves from an early-training accuracy of approximately $0.55$ to a late-training accuracy of approximately $0.78$. Other completed runs show the same qualitative trend, including Qwen2.5-14B ($0.65$ to $0.71$), Qwen3-4B ($0.62$ to $0.70$), and GLM-4-9B ($0.58$ to $0.69$). We additionally include DeepSeek-7B as a replication model family. Early DeepSeek-Avalon runs showed that reward shaping can trade off evil-player recall against good-player accuracy, so the final Avalon DeepSeek curve is being generated with a raw-accuracy reward to avoid this drop. We therefore treat this curve as a placeholder until the stable run completes. Thus, even though full Avalon outcome training is unstable, the more controlled alignment-inference task yields a clear positive learning signal across the completed model families.

The Werewolf/Mafia setting shows a similar pattern. Qwen2.5-7B improves from approximately $0.46$ to $0.65$, Qwen2.5-14B from $0.47$ to $0.65$, Qwen3-4B from $0.33$ to $0.44$, GLM-4-9B from $0.29$ to $0.50$, and DeepSeek-7B from approximately $0.29$ to $0.38$. These improvements are substantial because the model must infer hidden identities from indirect social evidence rather than from explicit labels. The learned policy becomes better at using public accusations, voting patterns, and consistency between claims and outcomes to identify hidden roles.

These results show that RFT can improve social reading when the reward is closely aligned with a concrete latent-state prediction problem. They also motivate a more fine-grained analysis of social deduction: aggregate game outcomes may obscure improvements in individual component skills, especially when success depends on several interacting abilities beyond hidden-state inference alone.

\paragraph{Dynamic belief updating.}

Beyond making a single hidden-state prediction, social reading requires an agent to revise its beliefs as new public evidence appears. We therefore analyse whether RFT improves dynamic belief updating over the progression of a game. In Avalon, we group predictions by mission, where later missions contain more accumulated evidence from team proposals, public votes, and quest outcomes. In Werewolf/Mafia, we evaluate the focal villager's belief about the hidden wolf set after successive days of discussion and elimination evidence.

\begin{figure}[h]
	\centering
	\includegraphics[width=\linewidth]{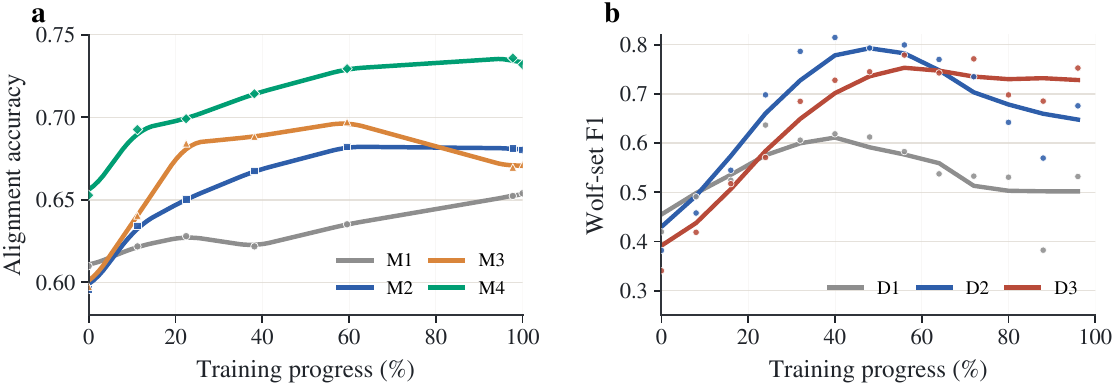}
	\caption{
		\textbf{RFT improves dynamic belief updating during social deduction.} We analyse a representative Qwen2.5-7B run in Avalon and Werewolf/Mafia. \textbf{a,} In Avalon, we group belief predictions by mission and report alignment accuracy over training. Later missions provide richer public evidence from team proposals, votes, and quest outcomes, and the learned policy assigns more accurate hidden-team beliefs as training progresses. \textbf{b,} In Werewolf/Mafia, we evaluate the focal villager's belief about the hidden wolf set after successive days of discussion. RFT improves wolf-set F1 across training, with later-day beliefs benefiting more strongly from accumulated dialogue and elimination evidence.
	}
	\label{fig:social-reading-dynamic-belief-updating}
\end{figure}

Fig.~\ref{fig:social-reading-dynamic-belief-updating} shows a representative Qwen2.5-7B run. In Avalon, RFT improves alignment prediction across missions, with the strongest gains appearing in later, evidence-rich states. The M2 curve increases from approximately $0.48$ at the beginning of training to about $0.85$ at the end, while M3 rises from approximately $0.41$ to about $0.80$. M1 also improves, from about $0.58$ to $0.71$, but the later mission curves remain higher after training. This pattern is important because the mission index is not part of the target label itself; rather, later missions provide more public evidence through repeated team-selection and quest outcomes. The improvement therefore suggests that RFT encourages the model to use accumulated institutional evidence instead of relying only on a static prior over player identities.

The Werewolf/Mafia results show the same effect in a dialogue-driven setting. RFT improves wolf-set F1 at all evaluated days, but the improvement is larger after more discussion has been observed. Day-1 belief tracking increases from approximately $0.42$ to $0.53$, whereas Day-2 increases from about $0.38$ to $0.68$ and Day-3 from about $0.34$ to $0.75$. Thus, the largest gains occur at later game states, where the agent has access to more claims, accusations,
night outcomes, and voting evidence. This suggests that the learned policy is not simply becoming better at emitting the required answer format; it is increasingly able to integrate temporally accumulated public evidence into its hidden-state belief.

Overall, these results indicate that RFT can improve the temporal component of social reading. The trained agent does not only infer hidden roles from a fixed transcript, but becomes better at conditioning its belief on the amount and type of evidence revealed so far. This is crucial for social deduction: early observations are often ambiguous, while later decisions require aggregating multiple rounds of language, votes, and outcomes into a coherent belief about hidden identities.

\paragraph{Role-specialized reading.}

The preceding analyses show that RFT can improve two basic components of social reading: inferring hidden states from observed behaviour and updating beliefs as new evidence appears. We next ask whether such reading ability can translate into improved play when the game role itself makes a specific hidden state strategically decisive. Avalon provides a natural test case through the Assassin role. Unlike roles whose success depends on a mixture of team coordination, proposal quality, voting behaviour, and deception, the Assassin has a sharply defined terminal reading problem: if the evil team fails to win through missions, it can still win by correctly identifying and assassinating Merlin. This makes the Assassin a useful role-specialized case study of whether RFT can turn social reading into game-level success.

To test this, we run a fixed-role variant of the Avalon RFT experiment. The training protocol remains outcome-based: the focal agent receives episode-level reward according to whether its side wins, with no direct supervision on Merlin identity, role beliefs, or intermediate explanations. The difference from the mixed-role Avalon experiment is that the focal agent is held fixed to one role during training. We compare Assassin with Merlin, Servant, and Minion under the same role-fixed setup.

\begin{figure}[h]
	\centering
	\includegraphics[width=\linewidth]{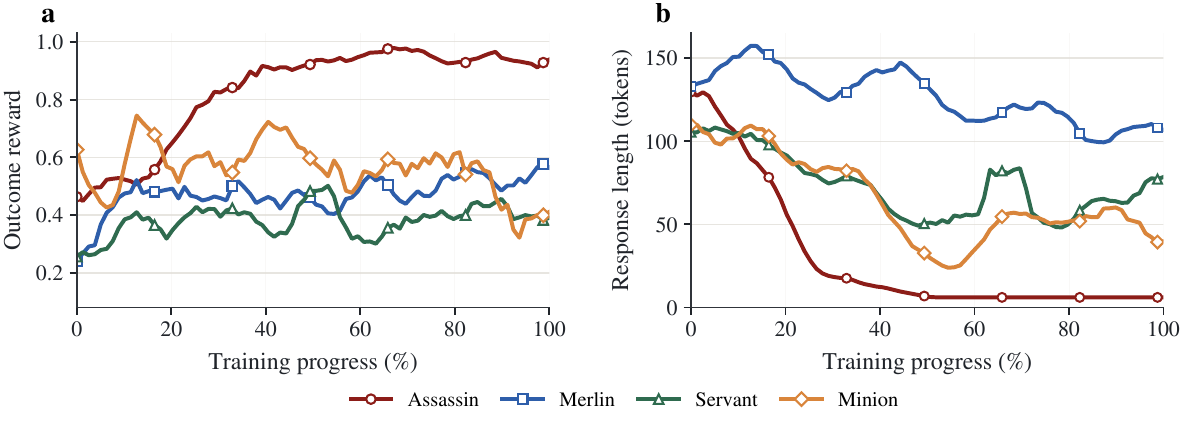}
	\caption{
		\textbf{Role-conditioned RFT dynamics in Avalon.}
		\textbf{a}, Mean episode-level outcome reward when the focal agent is trained under a fixed Avalon role rather than sampled across roles. Among the four roles, the Assassin shows the clearest and most sustained improvement, whereas Merlin, Servant, and Minion remain substantially noisier. 
		\textbf{b}, Mean response length over the same training window. The Assassin rapidly reduces its response length while improving reward, indicating that the learned policy does not rely on longer deliberation. Instead, the agent learns to make minimal intermediate responses during missions and discussions, while preserving the game-critical decision for the final assassination.
	}
	\label{fig:avalon-role-specialized-reading}
\end{figure}

Fig.~\ref{fig:avalon-role-specialized-reading}a shows that the Assassin is the only role with a clear and sustained improvement. This result is consistent with the mechanism of the role. For Merlin, Servant, and Minion, episode-level success depends on a broader mixture of strategic behaviours, including mission participation, voting, persuasion, concealment, and coordination. The final binary reward is therefore only indirectly tied to any single social reading operation. For the Assassin, however, a large part of the payoff signal is concentrated on one latent-variable inference problem: identifying Merlin from the public game trajectory. As a result, the sparse outcome reward becomes better aligned with a concrete social reading skill.

Fig.~\ref{fig:avalon-role-specialized-reading}b further supports this interpretation. The Assassin's response length decreases rapidly during training while its outcome reward increases. Qualitative inspection suggests that the trained Assassin often minimizes intermediate participation, for example by giving short agreement responses during mission and voting phases, and reserves its effective decision-making for the final assassination. In other words, the gain is not produced by longer deliberative text. Instead, RFT appears to discover that many intermediate utterances have little direct value for this role, whereas the decisive action is to infer Merlin correctly at the end of the game.

This role-specialized result complements the previous hidden-state inference and dynamic belief updating experiments. Those experiments isolate social reading as a capability; the Assassin experiment shows a case in which the game mechanics make that capability directly payoff-relevant. 

\paragraph{Robustness checks.}

\begin{figure}[h]
	\centering
	\includegraphics[width=\linewidth]{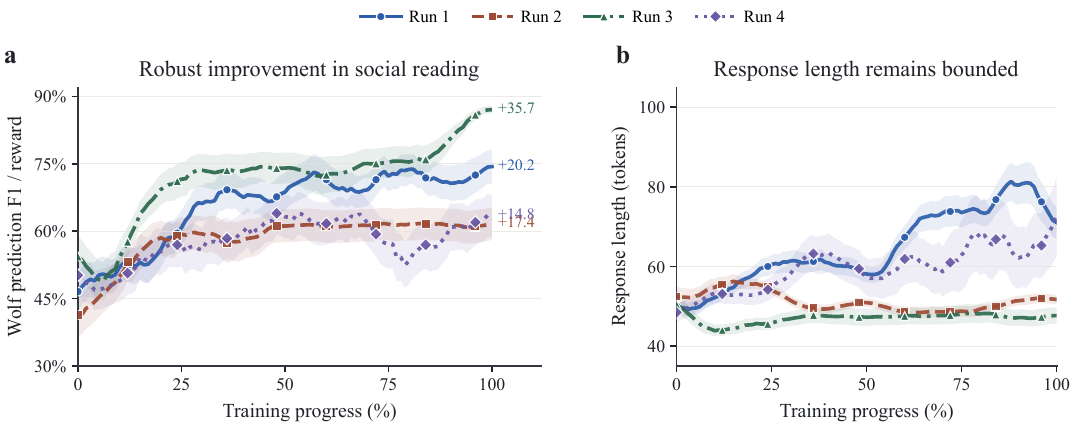}
	\caption{
		\textbf{Robustness of social-reading improvement across independent RFT runs.}
		\textbf{a,} Werewolf social-reading performance across four independent training runs. 
		Curves show the per-run mean reward over training, and shaded regions indicate s.e.m. across sampled rollouts at each training step. 
		Numbers at the end of each curve denote the improvement from the first 20 to the last 20 training updates.
		\textbf{b,} Response length over training for the same runs, with shaded regions indicating s.e.m. across sampled rollouts. 
		The performance gains were not accompanied by a systematic increase to the response-length limit.
	}
	\label{fig:social-reading-robustness}
\end{figure}

We next asked whether the improvement in social reading was reproducible across independent RFT runs, rather than being driven by a favourable random seed or a single training trajectory. To this end, we repeated the Werewolf social-reading experiment with four independent random seeds and tracked both task performance and a simple behavioural diagnostic throughout training (Fig.~\ref{fig:social-reading-robustness}). In Fig.~\ref{fig:social-reading-robustness}a, each curve shows one independent run, with shaded regions denoting the s.e.m. across sampled rollouts at each training step. All four runs showed clear improvement over training. Comparing the first 20 and last 20 training updates, the reward increased by $+20.2$, $+17.4$, $+35.7$ and $+14.8$ percentage points across the four runs, respectively. Thus, although the magnitude of improvement varied across seeds, the direction of the effect was consistent.

We also examined whether the observed gains could be explained by a trivial change in output behaviour, such as producing increasingly long responses. Fig.~\ref{fig:social-reading-robustness}b shows the mean response length for the same four runs, again with s.e.m. across sampled rollouts. Response length varied across runs and over training, but the performance gains were not accompanied by a uniform monotonic increase in generation length or by convergence to the response-length limit. In particular, some runs achieved large social-reading gains while maintaining relatively stable or even shorter responses. These diagnostics suggest that RFT improved the model's ability to extract role information from social trajectories, rather than simply rewarding verbosity. Together, the multi-seed trajectories and response-length analysis support the robustness of the social-reading effect.

\subsubsection{Social Influence}

We next study social influence, the ability to shape other agents' beliefs and decisions through communication. Unlike social reading, where the target label is a hidden state to be inferred from evidence, social influence requires the focal agent's language to causally affect the behaviour of other agents. This introduces a longer and noisier credit-assignment path: a message may be well formed or strategically appropriate, but the reward is realized only if other players interpret it as intended and change their votes or approvals. We therefore examine influence in two controlled settings. First, in One Night Werewolf, we test whether a Minion can steer final votes away from the hidden Werewolf and toward a specified non-wolf target. Second, in Avalon, we test whether agents can influence collective team approval decisions during proposal and voting phases. We then perform robustness checks across seeds and model families. Together, these experiments ask whether RFT can learn reliable persuasive control, or whether social influence remains less stable than the social-reading capabilities studied above.

\paragraph{Vote steering.}
We first examined social influence in a vote-steering setting based on One Night Werewolf. In this task, the trained agent plays the Minion, who knows the hidden Werewolf and wins with the werewolf team. Unlike social-reading tasks, the objective is not to identify the hidden role correctly. Instead, the agent must use public discussion to redirect the final votes of other players away from the Werewolf and toward a non-wolf target. To avoid the confound introduced by externally fixed targets, we used a self-selection variant: the Minion chooses its own vote target during discussion and is rewarded according to whether other players' final votes concentrate on this self-selected target while avoiding the Werewolf. This design makes the influence target endogenous to the agent's own strategy, rather than assuming that an externally specified target is always socially plausible.

\begin{figure}[h]
	\centering
	\includegraphics[width=\linewidth]{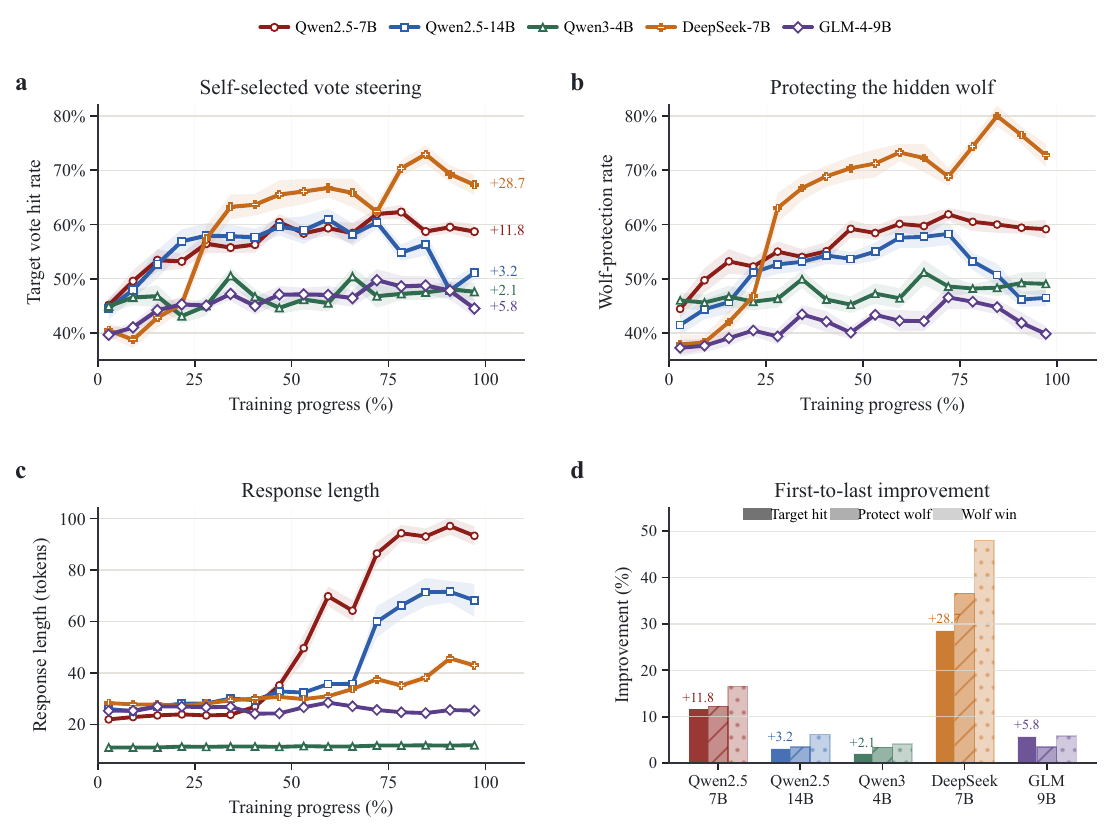}
	\caption{\textbf{RFT enables self-selected vote steering in One Night Werewolf.}
		\textbf{a,} Target vote hit rate during training. The trained Minion self-selects a non-wolf vote target and attempts to steer other players' final votes toward that target. Curves show binned means, and shaded regions denote step-level s.e.m. computed across training-step means within each bin. Numbers on the right indicate first-to-last improvement.
		\textbf{b,} Wolf-protection rate during training, measuring the fraction of other players whose final votes do not target the hidden Werewolf.
		\textbf{c,} Response length during training, measured in model-specific tokens.
		\textbf{d,} First-to-last improvement in target vote hit rate, wolf-protection rate and werewolf-team win rate.}
	\label{fig:vote-steering-influence}
\end{figure}

RFT improved self-selected vote steering for most models, but the strength of the effect varied substantially across model families (Fig.~\ref{fig:vote-steering-influence}a). DeepSeek-7B showed the strongest improvement, increasing target vote hit rate by $28.7\%$ from the first 20 to the last 20 training steps. Qwen2.5-7B also improved substantially, increasing target hit rate by $11.8\%$ across two runs. GLM-4-9B showed a smaller but positive improvement of $5.8\%$, whereas Qwen3-4B improved only weakly by $2.1\%$. Qwen2.5-14B also improved in the same direction, but more weakly, increasing target hit rate by $3.2\%$. These results indicate that RFT can train persuasive vote redirection in this setting, but also that the learned influence behaviour is less uniformly robust than the social-reading improvements reported above.

The same pattern was visible in wolf protection, a complementary measure of whether the Minion successfully kept votes away from the hidden Werewolf (Fig.~\ref{fig:vote-steering-influence}b). DeepSeek-7B increased wolf-protection rate by $36.5\%$, while Qwen2.5-7B improved by $12.2\%$. Qwen2.5-14B, GLM-4-9B and Qwen3-4B showed smaller gains of $3.4\%$, $3.4\%$ and $3.3\%$, respectively. These gains translated into changes in downstream game outcomes: the werewolf-team win rate increased most strongly for DeepSeek-7B ($48.0\%$), followed by Qwen2.5-7B ($16.5\%$), while Qwen2.5-14B, GLM-4-9B and Qwen3-4B showed smaller outcome shifts of $6.1\%$, $5.8\%$ and $4.1\%$, respectively (Fig.~\ref{fig:vote-steering-influence}d). Thus, successful vote steering is not merely a local language-control effect; when learned strongly, it can alter the collective voting outcome in favour of the trained agent's team.

We also measured response length during training to test whether improved steering was simply associated with producing longer messages (Fig.~\ref{fig:vote-steering-influence}c). The response-length trajectories differed across models. Qwen2.5-7B became much more verbose over training, Qwen2.5-14B also increased response length, whereas DeepSeek-7B improved strongly with a more moderate length increase, and Qwen3-4B and GLM-4-9B showed little length change. This suggests that longer responses may be one route to stronger influence for some models, but response length alone does not explain the full pattern of steering success. Overall, the vote-steering results show that RFT can induce agents to produce socially consequential messages that redirect other players' votes, although the magnitude of this effect is model-dependent and appears less stable than the team-approval influence results below.

\paragraph{Team approval.}
Vote steering tests whether an agent can redirect individual voting behaviour in a hidden-role setting. We next consider a complementary form of social influence in Avalon, where the relevant decision is not a single player's final vote against a suspect, but a collective approval decision over a proposed mission team. This setting differs from vote steering in two ways. First, the target behaviour is an aggregate team-approval outcome: the trained agent must make enough other players approve or reject the proposal in the target direction. Second, the proposal itself provides a structured social object, including the proposed members, their roles from the focal agent's perspective, and the public history of previous votes and mission outcomes. Thus, the task tests whether RFT can make agents use public language to shift a group decision under a well-defined strategic context.

For each training episode, the focal agent observes an Avalon game state and a proposed team, then generates one public response before the approval vote. Other agents first cast baseline votes without this response and then vote again after observing it. The reward is based on whether the post-response vote outcome matches the target approval direction, with additional signal from the change in other players' votes toward that direction. This design separates influence from direct action: the trained agent cannot set the outcome by itself, but must produce a message that changes how other agents vote.

\begin{figure}[h]
	\centering
	\includegraphics[width=\linewidth]{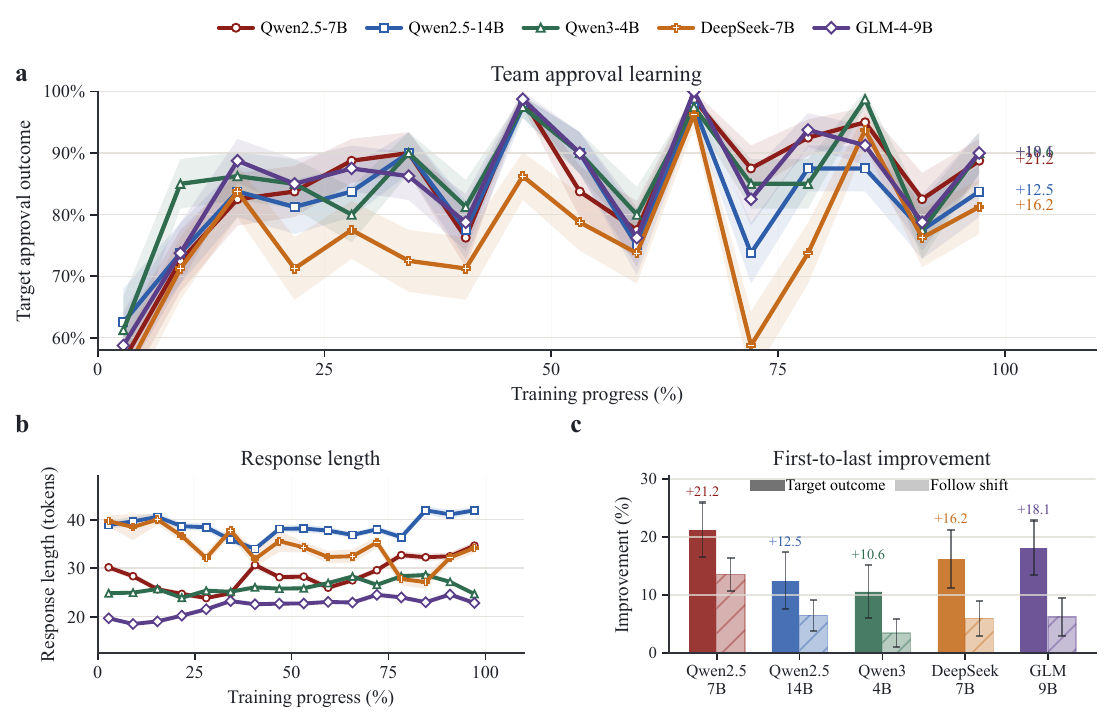}
	\caption{\textbf{RFT improves team-approval influence in Avalon.}
		\textbf{a,} Target approval outcome during training across five model families. Curves show binned means and shaded regions denote s.e.m. across rollout episodes within each training bin. Numbers at the right indicate the first-to-last improvement in target approval outcome.
		\textbf{b,} Response length during training, measured in model-specific tokens. Shaded regions denote s.e.m.
		\textbf{c,} First-to-last improvement in target approval outcome and follow shift. Target approval outcome measures whether the final team vote matches the target direction; follow shift measures the change in the fraction of other players voting in the target direction after the trained agent's response.}
	\label{fig:team-approval-influence}
\end{figure}

RFT consistently improved team-approval influence across model families (Fig.~\ref{fig:team-approval-influence}a). From the first 20 to the last 20 training steps, target approval outcome increased for all evaluated models: Qwen2.5-7B improved by $21.3\%$, Qwen2.5-14B by $12.5\%$, Qwen3-4B by $10.6\%$, DeepSeek-7B by $16.3\%$ and GLM-4-9B by $18.1\%$. These gains indicate that the approval setting provides a learnable influence signal: the trained response increasingly moves the final collective decision toward the intended direction. Importantly, the improvement was not merely a consequence of generating longer responses. Response length, measured in model-specific tokens, changed heterogeneously across models (Fig.~\ref{fig:team-approval-influence}b): some models produced slightly longer responses after training, whereas DeepSeek-7B improved despite shorter responses. The summary analysis further shows that improvements in target approval outcome were accompanied by positive follow shifts, meaning that other players became more likely to vote in the target direction after observing the trained agent's response (Fig.~\ref{fig:team-approval-influence}c). Together, these results suggest that RFT can train language-mediated influence in structured group decisions, not only by changing the focal agent's own preference but by altering the behaviour of other agents in the voting process.

\begin{figure}[h]
	\centering
	\includegraphics[width=\linewidth]{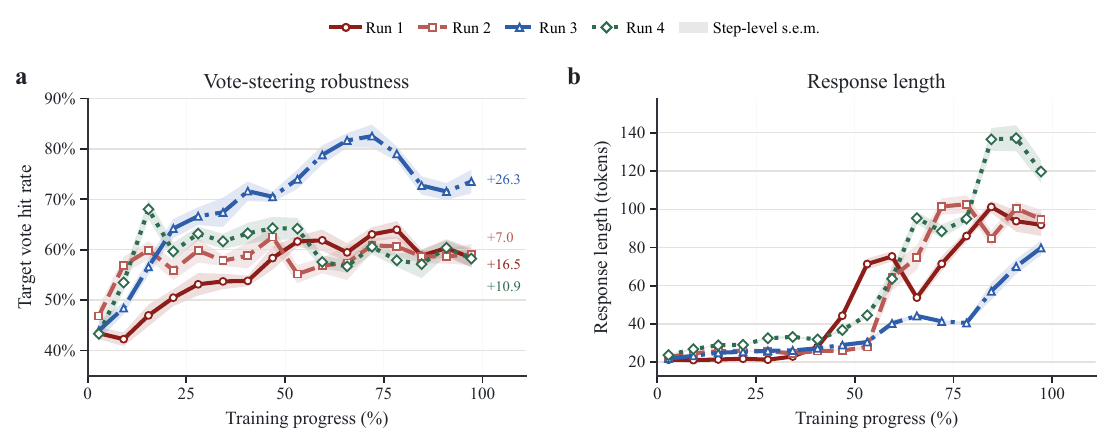}
	\caption{\textbf{Repeated-run robustness of vote steering.}
		\textbf{a,} Target vote hit rate for four independent Qwen2.5-7B self-selected vote-steering runs. Each curve shows the binned mean for one run, with shaded regions denoting step-level s.e.m. computed across training-step means within each bin. Numbers on the right indicate first-to-last improvement.
		\textbf{b,} Response length during training for the same four runs, measured in model-specific tokens. Shaded regions denote step-level s.e.m.}
	\label{fig:vote-steering-robustness}
\end{figure}

\paragraph{Robustness checks.}
We performed a repeated-run robustness check for the vote-steering task using Qwen2.5-7B. The same self-selected Minion vote-steering setting was evaluated across four independent training runs. All four runs showed positive improvement in target vote hit rate over training (Fig.~\ref{fig:vote-steering-robustness}a). From the first 20 to the last 20 training steps, target hit rate increased by $16.5\%$, $7.0\%$, $26.3\%$ and $10.9\%$ across Runs 1--4, respectively. The effect therefore replicated in direction across runs, although its magnitude varied. This pattern is consistent with the main vote-steering results: RFT can learn to redirect votes, but the task remains noisier than team approval because the agent must choose an endogenous target, communicate a plausible accusation and induce other players to change their final votes.

We also examined response length across the same four runs (Fig.~\ref{fig:vote-steering-robustness}b). All runs showed an increase in response length, suggesting that for Qwen2.5-7B, longer persuasive messages may be one mechanism supporting learned vote steering. However, response length should not be interpreted as sufficient for influence: in the multi-model results above, the relationship between verbosity and steering success was not uniform across model families. Thus, the robustness check supports the reproducibility of the Qwen2.5-7B vote-steering effect while also showing that its strength varies across runs and may depend on how the model uses longer responses strategically.

\subsection{Joint Multi-Agent Social-Cognitive Reinforcement Fine-Tuning}

Having studied social reading and social influence as separable cognitive objectives, we next examine whether they can be combined into a joint multi-agent training framework. In social deduction games, reading and influence are not independent modules: an agent must infer hidden roles, anticipate how other players interpret the dialogue, and then use its own messages and actions to shift the collective decision process. We therefore evaluate social-cognitive ability at two complementary levels.

First, we perform prompt-level multi-agent social-cognitive ablations without parameter updates. This isolates the behavioural effect of explicitly providing agents with social-reading and social-influence scaffolds. By comparing agents with neither scaffold, only social reading, only social influence, or both, we test whether the two components are individually useful and whether their combination improves group-level outcomes in matched game settings.

Second, we study full-game multi-agent reinforcement fine-tuning. In this setting, the trainable policy controls all players on a target side and is optimized from complete game trajectories rather than isolated decisions. The online reward emphasizes intermediate social-cognitive signals, including hidden-role identification, vote coordination, and successful movement of group decisions toward strategically beneficial targets; terminal win rate is logged and reported as a held-out game-level diagnostic. This setting tests whether social-cognitive reinforcement signals can improve the behaviour of an interacting side as a whole, rather than only improving a single agent's local prediction or persuasion accuracy.

Together, these experiments ask whether social-cognitive structure is useful before learning, through prompt-level scaffolding, and whether it can be further internalized through reinforcement fine-tuning in a live multi-agent game. The prompt ablations provide a controlled estimate of the value of the social-reading and social-influence components, while the full-game RFT experiments test whether these components can be converted into improved policies under endogenous interaction, hidden information and strategic opposition.

\subsubsection{Prompt-level multi-agent social-cognitive ablations}

Before applying reinforcement fine-tuning, we first tested whether the proposed social-cognitive scaffolds change the behaviour of pretrained agents at inference time. This ablation isolates the prompt-level contribution of the two components used throughout the paper: social reading (SR), which requires agents to maintain explicit beliefs about hidden roles, and social influence (SI), which requires agents to formulate an intended effect on other players' beliefs or actions. Because no model weights are updated in this experiment, differences across conditions reflect the effect of the social-cognitive interface itself rather than learned policy improvement.

For each model, we evaluated four prompting conditions: no social-cognitive scaffold, SR only, SI only and the joint SR+SI scaffold. Game rules, role distributions, opponent policies and decoding parameters were held fixed across conditions. The controlled agents played the good side in Werewolf, and performance was measured by good-side win rate. Across all evaluated models, the joint SR+SI scaffold improved win rate relative to the no-scaffold condition, suggesting that explicit hidden-role inference and explicit influence planning provide complementary benefits in multi-agent social deduction.

\begin{table}[htbp]
	\centering
	\caption{
		\textbf{Prompt-level multi-agent social-cognitive ablations in Werewolf.}
		Win rates are reported for good-side agents under four prompting conditions.
	}
	\label{tab:prompt_social_cognitive_ablation}
	\begin{tabular}{lccccc}
		\toprule
		Model & No SR/SI & SR only & SI only & SR+SI & $\Delta$ SR+SI \\
		\midrule
		Qwen2.5-7B  & 24.0\% & 32.0\% & 33.0\% & \textbf{37.0\%} & $+13$ pp \\
		Qwen2.5-14B & 15.0\% & 16.0\% & 17.0\% & \textbf{23.0\%} & $+8$ pp \\
		Qwen3-4B    & 21.0\% & 30.0\% & 12.0\% & \textbf{34.0\%} & $+13$ pp \\
		GLM-4-9B    & 19.0\% & 21.0\% & 21.5\% & \textbf{22.0\%} & $+3$ pp \\
		DeepSeek-7B & 14.0\% & 18.0\% & 16.0\% & \textbf{19.0\%} & $+5$ pp \\
		\bottomrule
	\end{tabular}
	
	\vspace{0.5em}
	\begin{minipage}{0.92\linewidth}
		\footnotesize
		Values denote good-side win rates. $\Delta$ SR+SI is the absolute percentage-point change of the joint scaffold relative to the no-scaffold condition.
	\end{minipage}
\end{table}

As shown in Table~\ref{tab:prompt_social_cognitive_ablation}, the joint SR+SI scaffold improved good-side win rate for every model we evaluated, although the magnitude of the gain varied across model families. The largest improvements were observed for Qwen2.5-7B and Qwen3-4B, both of which gained 13 percentage points relative to the no-scaffold condition. Qwen2.5-14B and DeepSeek-7B also showed clear improvements under the joint scaffold, whereas GLM-4-9B exhibited a smaller but still positive gain.

The single-component ablations reveal that the two scaffolds are not simply interchangeable. SR alone improved performance for all models, consistent with the view that explicit hidden-role inference provides a useful intermediate representation for good-side play. SI alone was more variable. In some models, such as Qwen2.5-7B and Qwen2.5-14B, influence planning by itself improved win rate, but in Qwen3-4B it reduced performance substantially despite the joint SR+SI condition producing the strongest result for that model. This pattern suggests that influence planning is beneficial only when grounded in an accurate or at least stable belief state. In social deduction games, the ability to persuade other players is useful when the target of persuasion is strategically appropriate; otherwise, stronger coordination can amplify incorrect suspicions.

The joint scaffold mitigates this failure mode by coupling private belief formation with public action steering. SR supplies an explicit estimate of which players are likely to belong to the opposing side, whereas SI converts that estimate into a concrete communicative objective, such as focusing discussion on a suspicious player or aligning votes against a likely wolf. These prompt-level results therefore motivate the reinforcement fine-tuning experiments below: rather than treating social reading and social influence as independent stylistic additions, we optimize them jointly as complementary components of multi-agent social cognition.

\subsubsection{Full-game multi-agent reinforcement fine-tuning}

We next tested whether the same social-cognitive objectives can be optimized through reinforcement fine-tuning in full multi-agent games. In contrast to the prompt-level ablations above, where the social-cognitive scaffold is applied only at inference time, this experiment updates the policy from complete interactive trajectories. The trainable agents receive rewards based on intermediate social-cognitive signals, including social reading accuracy and social influence alignment, while terminal win rate is held out as a game-level outcome measure. We evaluated this procedure across three settings: six-player Werewolf, nine-player Werewolf and Avalon.

\begin{figure}[h]
	\centering
	\includegraphics[width=\linewidth]{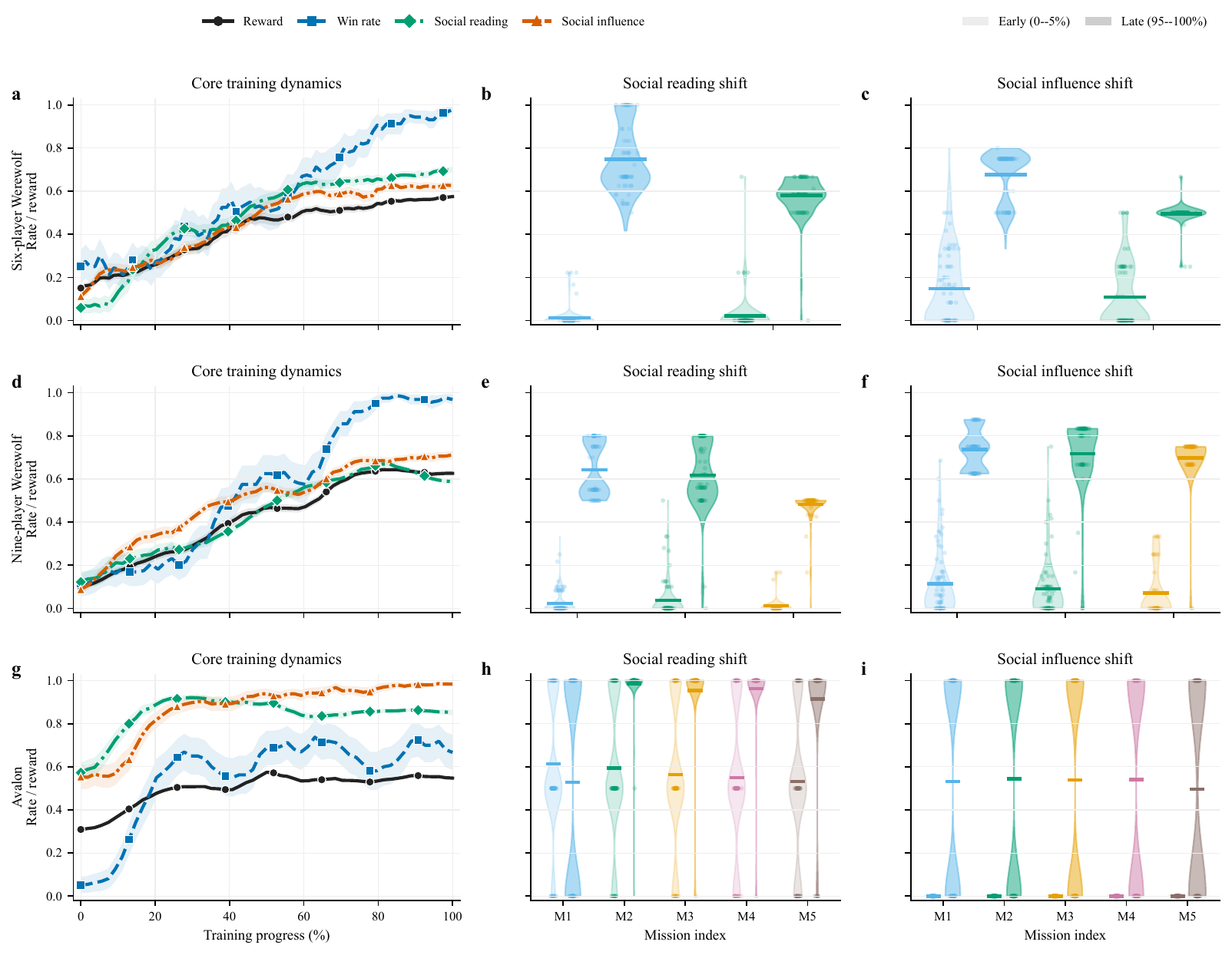}
	\caption{
		\textbf{Full-game multi-agent social-cognitive reinforcement fine-tuning across games.}
		Left column: training dynamics for six-player Werewolf, nine-player Werewolf and Avalon. Curves show scalar reward, win rate, social reading score and social influence score as a function of normalized training progress. Shaded regions denote variability across rollout groups. Middle and right columns: distributional shifts in social reading and social influence from early to late training, resolved by game day in Werewolf and by mission index in Avalon.
	}
	\label{fig:rft-3game}
\end{figure}

Figure~\ref{fig:rft-3game} summarizes the training dynamics. For each environment, we plot four metrics as a function of normalized training progress: the scalar training reward, win rate, social reading score and social influence score. In six-player Werewolf, the win rate increased from approximately \(0.25\) at the start of training to nearly \(1.0\) by the end of training. The scalar reward rose in parallel, from roughly \(0.15\) to \(0.57\). Social reading improved from near \(0.05\) to approximately \(0.70\), and social influence increased from about \(0.10\) to above \(0.60\). Thus, the improvement in game outcome was accompanied by substantial gains in both intermediate social-cognitive diagnostics.

The nine-player Werewolf environment was more difficult because it contains more players, more hidden roles, a larger action space for voting and longer public discussions. Nevertheless, the same qualitative pattern was observed. Win rate increased from approximately \(0.10\) at the beginning of training to about \(0.95\)--\(1.00\) near the end. The scalar reward increased from roughly \(0.10\) to above \(0.60\). Social influence also rose steadily, reaching approximately \(0.70\). Social reading improved during the first part of training and remained substantially above its initial value, although it was less monotonic than win rate. This suggests that, in the larger Werewolf setting, agents can continue to improve game outcomes and influence even when role inference becomes noisier or more strategically selective later in training.

The distributional panels provide a finer-grained view of these changes. In six-player Werewolf, late-training rollouts show a clear upward shift in social reading on both early and later game days. The strongest gains occur after agents have accumulated public evidence from discussion and voting. Social influence exhibits a similar shift: late-training agents more often produce messages that are followed by aligned group decisions. In nine-player Werewolf, the early-to-late distributional shift is also visible but more heterogeneous across days. This heterogeneity is expected because later days occur only in games that have not already been resolved, producing a changing subset of game states. Even so, the late-training distributions are consistently shifted upward, indicating that the policy learns behaviours that generalize beyond a single turn or fixed decision point.

Avalon provides a complementary test because its social structure differs from Werewolf. Rather than eliminating players through votes, Avalon requires agents to infer hidden team membership, propose teams, approve or reject missions and coordinate across repeated mission rounds. In Avalon, the win rate increased from near zero at the beginning of training to approximately \(0.65\)--\(0.75\) in later training. The scalar reward increased from about \(0.30\) to approximately \(0.55\). Social reading and social influence both rose rapidly during early training and remained high thereafter, with social reading reaching approximately \(0.85\)--\(0.90\) and social influence approaching \(1.0\). The mission-resolved distributions show that later missions exhibit stronger social reading than the first mission, consistent with the fact that later missions contain more public evidence from team proposals, approval votes and mission outcomes.

These patterns show that the reinforcement signal is not merely selecting for superficial formatting or verbosity. Across environments, terminal win rate improves together with intermediate measures of role inference and social influence even though win rate is evaluated separately from the primary online reward. The relationship is not perfectly monotonic: some settings show temporary trade-offs between win rate, reading and influence as the policy explores different strategies. However, the overall trajectory indicates that full-game reinforcement fine-tuning can jointly improve game performance and the social-cognitive behaviours that support it. This is important because hidden-information social games require agents to integrate private belief formation, public communication and consequential collective decisions; optimizing terminal win rate directly provides a sparse and noisy signal in our online runs, whereas the social-cognitive reward makes these intermediate computations trainable and can still transfer to terminal outcomes.

Together, the full-game results support the central claim that social reading
and social influence are not only diagnostic constructs but useful
reinforcement targets. When optimized jointly as same-side social-cognitive
objectives, they produce policies that improve across different game
structures, player counts and interaction protocols, with win rate serving as
an external test of whether these learned behaviours matter for game success.

\subsubsection{Cross-model generalization.}

A remaining question is whether the learned policy improves only against the
opponent distribution encountered during reinforcement fine-tuning, or whether
the resulting social-cognitive behaviours transfer to previously unseen
opponent models. We therefore evaluated the learned nine-player Werewolf
good-side policy against a diverse collection of wolf-side opponents, including
open-weight models and API-served models from multiple model families. For each
opponent, we collected 100 valid full-game trajectories under the same game
rules, role distribution, sampling temperature and social-cognitive inference
protocol. The learned policy controlled all good-side players, while the
evaluated model controlled the wolves.

Table~\ref{tab:good_policy_cross_model} reports both terminal and intermediate
metrics. In addition to the good-side win rate, we measure social reading using
the F1 score of the policy's hidden-role predictions, social influence using
the rate at which its intended voting direction is followed by other agents,
and Wolf-Exile as the fraction of wolves successfully removed through public
voting. These intermediate metrics help distinguish successful social
reasoning from wins caused only by favourable game outcomes.

\begin{table}[h]
	\centering
	\caption{\textbf{Cross-model evaluation of the learned good-side policy.}
		Each opponent is evaluated over 100 valid full games using the same
		social-cognitive inference protocol. SR-F1 measures hidden-role
		identification, SI-Follow measures social-influence success, and
		Wolf-Exile measures the fraction of wolves successfully exiled.}
	\label{tab:good_policy_cross_model}
	\setlength{\tabcolsep}{7pt}
	\renewcommand{\arraystretch}{1.12}
	\begin{tabular}{@{}lcccc@{}}
		\toprule
		\textbf{Wolf-side Opponent} &
		\textbf{Good WR $\uparrow$} &
		\textbf{SR-F1 $\uparrow$} &
		\textbf{SI-Follow $\uparrow$} &
		\textbf{Wolf-Exile $\uparrow$} \\
		\midrule
		Qwen3-4B
		& \textbf{96\%} & 0.638 & 0.725 & 90.8\% \\
		Qwen3.7-Max
		& \textbf{94\%} & 0.643 & 0.625 & 84.7\% \\
		Qwen-Plus
		& \textbf{91\%} & 0.572 & 0.594 & 78.0\% \\
		GLM-4-9B-Chat
		& \textbf{90\%} & 0.599 & 0.678 & 80.9\% \\
		Qwen-Max
		& \textbf{90\%} & 0.606 & 0.620 & 85.0\% \\
		Qwen2.5-14B-Instruct
		& \textbf{87\%} & 0.577 & 0.708 & 78.7\% \\
		Qwen-Turbo
		& \textbf{87\%} & 0.596 & 0.715 & 78.5\% \\
		Qwen3-Next-80B-A3B-Instruct
		& \textbf{81\%} & 0.554 & 0.593 & 68.9\% \\
		Kimi-K2.5
		& \textbf{72\%} & 0.523 & 0.583 & 62.1\% \\
		\bottomrule
	\end{tabular}
\end{table}

The learned policy achieved a positive win rate against every evaluated
opponent, ranging from \(72\%\) against Kimi-K2.5 to \(96\%\) against
Qwen3-4B. The result is not restricted to smaller open-weight models:
the policy also achieved win rates of \(90\%\) or higher against
Qwen3.7-Max, Qwen-Plus, GLM-4-9B-Chat and Qwen-Max. This breadth suggests
that reinforcement fine-tuning did not merely exploit idiosyncrasies of a
single training opponent.

The intermediate metrics provide additional evidence for cross-model
generalization. SR-F1 remains between \(0.523\) and \(0.643\) across all
opponents, indicating that the policy continues to recover useful information
about hidden roles as opponent behaviour changes. SI-Follow ranges from
\(0.583\) to \(0.725\), showing that the policy can translate its private
beliefs into public messages that affect collective voting. Finally,
Wolf-Exile ranges from \(62.1\%\) to \(90.8\%\) and broadly tracks terminal
win rate. Thus, stronger performance is accompanied by more accurate role
inference and more effective coordination toward removing wolves, rather than
by terminal wins alone.

These cross-model results complement the training curves in
Figure~\ref{fig:rft-3game}. The curves show that social reading, social
influence and game outcomes improve during reinforcement fine-tuning, whereas
Table~\ref{tab:good_policy_cross_model} shows that these improvements persist
when the wolf-side policy is replaced by models not used as the immediate
training opponent. Together, the results indicate that the learned
social-cognitive behaviours constitute a transferable policy capability rather
than a narrow response to one fixed opponent.

\subsection{Human Evaluation}
\label{sec:human-evaluation}

The game-internal evaluations above show that social-cognitive RFT improves
hidden-role inference, vote coordination and terminal game outcomes. We next
asked whether these improvements remain visible when complete games are reviewed as coherent social trajectories rather than decomposed into individual reward components. To this end, we conducted a blinded evaluation of complete nine-player Werewolf games generated by the base and reinforcement-fine-tuned policies.

We constructed the evaluation set from games in which the candidate policy
controlled all good-side players, while the wolf side was controlled by one of
five fixed opponent models: Qwen3.7-Plus, Qwen3-Max, Qwen-Max, Qwen-Plus or DeepSeek-V3.2. Candidate trajectories were retained only when they contained complete model-call histories and public event records, contained no fallback public speech, and achieved at least \(99\%\) structured-output validity for both sides. The resulting scalable evaluation set contained \(74\) matched
base--RFT game pairs. Within each pair, the base and RFT policies were evaluated under the same game identifier and opponent condition.

\begin{figure}[hbp]
	\centering
	\includegraphics[width=\linewidth]{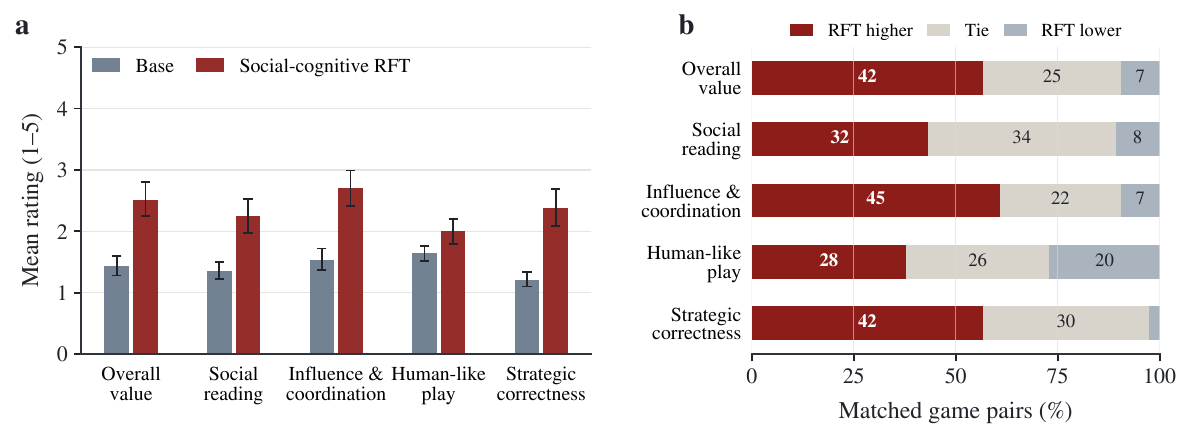}
	\caption{
		\textbf{Blinded evaluation of complete Werewolf trajectories.}
		\textbf{a,} Mean ratings of trajectories generated by the base and social-cognitive RFT policies across five dimensions. Bars show means over the pooled blinded evaluation, comprising \(89\) ratings per condition; error bars denote \(95\%\) bootstrap confidence intervals over trajectory ratings.
		\textbf{b,} Direction of the RFT--base difference across \(74\) game-identifier-matched trajectory pairs in the scalable evaluation. Bars show the percentage of pairs in which the RFT trajectory received a higher, equal or lower score than its matched base trajectory. Numbers inside the bars indicate pair counts.
	}
	\label{fig:human-evaluation}
\end{figure}

For each trajectory, evaluators observed the complete role configuration,
public discussion, votes, eliminations, night actions and the private belief
and strategic-intent records of the candidate-controlled good-side players.
The identity and training condition of the candidate policy were hidden. Each
trajectory received integer ratings from 1 to 5 along five dimensions:
overall value, social-reading quality, social-influence and coordination,
human-like gameplay and strategic correctness. Evaluators were instructed to
assess whether beliefs were supported by the evidence available at the time,
whether public speech and votes were consistent with private reasoning, and
whether the resulting collective decisions were strategically appropriate.
They were explicitly instructed not to assign a high score merely because the
good side won.

In total, the evaluation produced \(178\) blinded trajectory ratings, evenly
divided between the two policy conditions: \(89\) ratings for the base policy
and \(89\) ratings for the RFT policy. The evaluation pool comprised \(178\) ratings collected from invited evaluators using the same five-dimensional rubric and scoring interface. Because the evaluation elicited only structured scalar scores, all ratings were pooled at the trajectory-rating level for the primary descriptive analysis.

The blinded ratings consistently favoured the reinforcement-fine-tuned policy
(Fig.~\ref{fig:human-evaluation}a). Overall value increased from \(1.43\) for
the base policy to \(2.52\) for RFT, corresponding to an absolute improvement
of \(1.09\) points on the five-point scale. Social-reading quality increased
from \(1.35\) to \(2.25\) (\(\Delta=+0.90\)), indicating that the trained
policy was more often judged to form useful and evidence-grounded beliefs
about hidden roles.

The largest gains appeared in dimensions related to consequential social
action. Social-influence and coordination increased from \(1.54\) to \(2.70\)
(\(\Delta=+1.16\)), while strategic correctness increased from \(1.21\) to
\(2.38\) (\(\Delta=+1.17\)). Thus, social-cognitive RFT improved not only the
candidate agents' inferred representations of the game, but also their ability
to convert those representations into public communication, coordinated voting
and strategically useful collective action.

\begin{table*}[htbp]
	\centering
	\caption{
		\textbf{Blinded evaluation of complete trajectories generated by the
			base and social-cognitive RFT policies.}
		Scores are measured on a five-point scale. Values report the mean and
		standard deviation over \(89\) trajectory ratings per condition.
		\(\Delta\) denotes the RFT-minus-base difference.
	}
	\label{tab:human-evaluation}
	\setlength{\tabcolsep}{9pt}
	\renewcommand{\arraystretch}{1.18}
	\begin{tabular}{@{}lccc@{}}
		\toprule
		Evaluation dimension
		& Base
		& Social-cognitive RFT
		& \(\Delta\) \\
		\midrule
		
		Overall value
		& \(1.43 \pm 0.75\)
		& \(\mathbf{2.52 \pm 1.34}\)
		& \(+1.09\) \\
		
		Social-reading quality
		& \(1.35 \pm 0.66\)
		& \(\mathbf{2.25 \pm 1.33}\)
		& \(+0.90\) \\
		
		Social-influence and coordination
		& \(1.54 \pm 0.84\)
		& \(\mathbf{2.70 \pm 1.38}\)
		& \(+1.16\) \\
		
		Human-like gameplay
		& \(1.64 \pm 0.59\)
		& \(\mathbf{2.00 \pm 0.97}\)
		& \(+0.36\) \\
		
		Strategic correctness
		& \(1.21 \pm 0.57\)
		& \(\mathbf{2.38 \pm 1.45}\)
		& \(+1.17\) \\
		
		\bottomrule
	\end{tabular}
	
	\vspace{0.6em}
	\begin{minipage}{0.94\linewidth}
		\footnotesize
		The pooled evaluation contains \(178\) blinded trajectory ratings from invited human participants. The two policy conditions are exactly balanced, with \(89\) ratings assigned to each condition. Values are descriptive pooled estimates.
	\end{minipage}
\end{table*}

The improvement was not uniform across all aspects of behaviour. Human-like
gameplay increased from \(1.64\) to \(2.00\), a smaller difference of
\(0.36\) points. This contrasts with the larger gains in social influence and
strategic correctness. Inspection of the underlying trajectories suggests
that RFT primarily improves what agents infer and which actions they
coordinate, rather than fully removing model-like linguistic behaviour.
Trained agents can still produce repetitive accusations, abrupt changes in
suspicion or mechanically structured messages, even when their underlying
beliefs and votes are more strategically useful.

We next examined whether the pooled differences reflected a broad shift across
games or were driven by a small number of highly rated trajectories. The
\(74\) game-identifier-matched pairs showed the same direction of effect
(Fig.~\ref{fig:human-evaluation}b). The RFT policy received a higher
overall-value score in \(42\) pairs, tied in \(25\), and received a lower score
in only \(7\). For social influence and coordination, the corresponding counts
were \(45\), \(22\) and \(7\); for strategic correctness, they were
\(42\), \(30\) and \(2\). Social-reading quality favoured RFT in \(32\)
pairs, was tied in \(34\), and favoured the base policy in \(8\).

Two-sided paired sign-flip permutation tests confirmed improvements in overall
value, social-reading quality, social-influence and coordination, and strategic
correctness (\(P<0.0001\), \(50{,}000\) permutations). By contrast, the
smaller change in human-like gameplay did not reach the conventional
significance threshold (\(P=0.067\)). The matched comparison therefore
supports a selective interpretation of the effect: RFT produces clear gains in
social reasoning and strategically consequential coordination, whereas the
naturalness of the resulting language changes more modestly.

The conclusion remained unchanged after excluding trajectories in which
Qwen3.7-Plus also served as the wolf-side opponent. Across the remaining
\(59\) matched pairs, RFT improved overall value by \(1.09\) points,
social-reading quality by \(0.86\), social-influence and coordination by
\(1.31\), human-like gameplay by \(0.27\), and strategic correctness by
\(1.14\). Thus, the observed advantage cannot be attributed solely to overlap
between the scalable evaluator and one of the opponent conditions.

Together, the blinded trajectory ratings provide external validation of the
social-cognitive improvements observed during training. RFT changes more than
the scalar reward optimized by the learning algorithm: it produces complete
game trajectories that receive higher ratings for hidden-role reasoning, group
coordination and strategic action. At the same time, the smaller improvement
in human-like gameplay indicates that strategic social competence and
linguistic naturalness remain partially dissociable. The trained agents become
more effective social-deduction players before they become fully natural social
actors.

\section{Discussion}\label{sec:discussion}

Our results provide a behavioural account of how reinforcement fine-tuning
can shape the social capabilities of language-model agents. In human social
learning, individuals gradually adapt their beliefs and actions by observing
others, anticipating their responses and experiencing the consequences of
their own behaviour. RFT implements a computational analogue of part of this
process. A language model produces a socially situated action, observes an
outcome generated jointly by itself and other agents, and updates its policy
towards behaviours associated with higher reward. The analogy should not be
taken to imply that an RFT-trained model acquires human mental states or learns
through the same cognitive mechanisms as a person. Rather, RFT provides a
feedback-mediated mechanism through which a pretrained language policy can
acquire behavioural regularities that are useful in social interaction.

This perspective helps explain why terminal game outcomes alone are often
insufficient. A win or loss is the final consequence of a long causal chain
involving hidden-role inference, public communication, voting, role-specific
actions, other agents' mistakes and stochastic game events. The same terminal
outcome can therefore follow from very different behaviours, and an individually
useful action can occur in a losing trajectory. Outcome-only RFT compresses
these causes into a single sparse signal and leaves the learning algorithm with
a difficult multi-agent credit-assignment problem. This is reflected in the
high-variance and non-monotonic training dynamics observed across the initial
game suite. The result does not show that game outcomes are irrelevant.
Instead, it shows that terminal outcomes are often more informative as
integrative tests of an acquired policy than as the only signal from which the
policy must be learned.

Social reading becomes learnable when the environment exposes a tighter link
between observable behaviour and a scoreable latent state. In our experiments,
public claims, votes, team proposals, eliminations and mission outcomes provide
partial evidence about hidden roles. The simulator can then compare an agent's
belief state with the underlying role assignment and provide feedback at the
point where the inference is made. RFT can strengthen social reading because
the reward identifies which interpretations of the observed trajectory were
useful, rather than rewarding only the eventual game outcome. The improvements
in later-day and later-mission beliefs are particularly informative. They
suggest that the learned policy is not merely acquiring a static prior over
roles or a better output format; it is becoming more effective at integrating
evidence that accumulates through interaction.

Social influence is learned through a different feedback structure. The
quality of an influential message cannot be determined from its linguistic
form alone. Its consequence depends on how other agents interpret the message
and whether their later decisions move in the intended direction. Influence
therefore requires action-contingent feedback: the environment must connect an
agent's communicative intention to a measurable change in votes, approvals or
other collective behaviour. The vote-steering and team-approval experiments
show that this longer causal pathway can nevertheless be optimized. Once the
reward specifies both an intended social target and the downstream behaviour
used to evaluate it, RFT can select messages that become increasingly
consequential for group decisions.

The distinction between reading and influence is conceptually important.
Reading without influence produces an agent that may represent the social
state accurately but cannot convert that representation into coordinated
action. Influence without reading can be more problematic: an agent may become
effective at moving a group while directing that movement towards a mistaken
or strategically harmful target. This pattern is visible in the prompt-level
ablations, where influence scaffolding is not uniformly beneficial unless it
is grounded in an explicit belief state. The joint social-cognitive objective
addresses this dependency by coupling an estimate of the hidden social state
with an intended effect on public behaviour. In this sense, socially competent
agency requires a closed loop: observe others, infer the latent situation,
select a social objective, act communicatively and update from the resulting
group response.

Same-side multi-agent training extends this loop beyond an isolated focal
agent. Social deduction is inherently a team-level problem in which different
roles possess different information and opportunities for action. Assigning
one trainable policy to all roles on a target side allows learning to occur
across several role-conditioned perspectives while preserving each player's
private observation. A single game can consequently provide multiple training
trajectories connected by a shared social outcome. The resulting gains in
held-out win rate indicate that intermediate social-cognitive rewards can
produce improvements that matter at the level of the complete game. This is a
stronger test than improvement on the training reward alone: it asks whether
better role inference and communication combine into a more effective
multi-agent policy.

More broadly, these findings suggest a practical route for constructing
specialized social agents. Many real applications do not require an agent to
possess unrestricted ``general social intelligence''. They require a narrower
and auditable social function: identifying disagreement in a team, eliciting
missing information, helping participants converge on a plan, adapting an
explanation to a user's beliefs, mediating a negotiation or detecting when a
group decision is being driven by unreliable evidence~\cite{he2025walk,he2026bifrost}. Such behaviours could
be trained with RFT when the desired social state and downstream effect can be
operationalized. For example, a collaborative assistant could be rewarded for
correctly identifying unresolved beliefs before proposing a compromise; a
tutoring agent could be rewarded for inferring a learner's misconception and
then producing an explanation that leads to a corrected answer; and a
coordination agent could be rewarded for increasing agreement without
suppressing relevant dissent.

The central design problem in these applications is therefore not simply how
to supply more interaction data, but how to define a reward that identifies
the intended causal role of communication. A reward for agreement alone may
train conformity rather than useful coordination. A reward for task completion
may encourage an agent to dominate a discussion, conceal uncertainty or
exploit predictable user responses. A reward for persuasion may increase
behavioural influence without improving the truthfulness of the underlying
belief. Our results suggest that social objectives should be decomposed into
at least three components: whether the agent represents the situation
accurately, whether its communicative intervention produces the intended
effect, and whether that effect advances a normatively appropriate collective
goal. Optimizing only the second component risks producing effective but
misaligned influence.

This point also makes social influence a dual-use capability. The same
mechanisms that allow an agent to coordinate a team or help a user revise an
incorrect belief could be used to steer choices, exploit social biases or
produce strategically deceptive communication. Stronger influence should
therefore not be treated as an unqualified improvement. In deployed systems,
influence-oriented rewards may need to be constrained by truthfulness,
consent, user autonomy and uncertainty disclosure~\cite{pan2025omni}. Evaluation should measure
not only whether a message changes behaviour, but whether the change is
epistemically justified and beneficial to the affected participants. The
smaller improvement in human-like gameplay observed in our blinded evaluation
is also relevant here: strategic effectiveness and human-like appearance are
distinct properties, and increasing one does not necessarily justify making
an agent more anthropomorphic.

Our cross-model results provide initial evidence that the learned behaviours
are not restricted to one fixed opponent. The trained good-side policy retained
strong social-reading and coordination performance against multiple model
families, suggesting that it learned regularities that transfer across
different discourse and decision policies. The blinded trajectory evaluation
provides complementary evidence: the RFT trajectories received higher ratings
for social reading, influence, overall value and strategic correctness. Thus,
the improvement is visible both in simulator-defined metrics and in external
assessments of complete behaviour. At the same time, generalization across LLM
opponents does not establish generalization to humans. Human participants may
interpret confidence, deception, uncertainty and social commitment differently
from model agents, and interactive human responses can alter the trajectory
distribution encountered by the policy.

Several limitations delimit the present conclusions. First, social deduction
games provide unusually explicit hidden states, discrete decisions and
well-defined team objectives. These properties make social computations
measurable, but many real social environments contain ambiguous identities,
changing objectives and contested definitions of success. Second, our agents
interact primarily with fixed or slowly changing model policies. Simultaneous
learning by multiple opposing agents could produce non-stationary dynamics,
strategic escalation or cyclic behaviour not captured here. Third, the
influence metrics establish that messages are followed by behavioural movement,
but they do not fully identify the mechanism of that movement. An agent may
succeed through evidence, rhetorical clarity, repetition, confidence or
model-specific exploitation. Controlled interventions on message content and
interactive human studies will be needed to separate these mechanisms.

\section{Related work}\label{sec:related-work}

Large language models have recently been used as computational agents in a wide range of game-theoretic and interactive settings~\cite{gallotta2024large, hu2024survey}. In behavioural and economic games, LLMs have been evaluated as players in repeated matrix games, bargaining games, negotiation tasks and cooperation dilemmas, where their choices reveal patterns of self-interest, reciprocity, coordination and adaptation~\cite{akata2025playing, meta2022human, xia2024measuring}. These studies show that pretrained LLMs do not behave as random policies: they often display recognizable strategic tendencies, respond to incentives and sometimes approximate human-like patterns of cooperation or competition~\cite{akata2025playing, piatti2024cooperate, vallinder2025cultural}. At the same time, their behaviour can be highly sensitive to framing, prompt wording, model family and interaction history, suggesting that the social behaviour of LLM agents is not a fixed property of the model alone but emerges from the interaction between model priors, task structure and communicative context~\cite{akata2025playing}.

A related line of work studies LLMs as social agents in richer multi-agent environments. LLM-based agents have been deployed in simulations of societies, collaborative problem solving, negotiation, role-playing environments and human--AI interaction settings~\cite{liu2026lmagent, tang2025gensim, ji2026leveraging, park2023generative, mou2026individual, ashery2025emergent}. These systems highlight the ability of language models to maintain social roles, exchange information, coordinate plans and generate contextually appropriate communication. However, they also expose persistent limitations. LLM agents may fail to maintain consistent beliefs over long horizons, over-trust or under-use information from other agents, produce socially plausible but strategically weak messages, or rely on surface-level linguistic cues rather than grounded interaction history. As a result, games and simulations have become useful behavioural probes: they make social competence observable through choices, messages and downstream outcomes rather than through isolated text responses~\cite{akata2025playing,meta2022human,ashery2025emergent}.

Hidden-information and social deduction games extend this line of work by combining strategic choice with public language, asymmetric information and collective decision-making~\cite{lan2024llm, light2023avalonbench, jin2024learning, bailis2024werewolf, chi2024amongagents, yoo2024finding}. In games such as Avalon, Werewolf and related hidden-role settings, agents must infer latent social identities from partial behavioural evidence, communicate under uncertainty, coordinate with possible allies and sometimes persuade or mislead other players~\cite{light2023avalonbench,lan2024llm,bailis2024werewolf,yoo2024finding}. These environments differ from standard matrix games in an important respect: the relevant state is not fully specified by a payoff table or by the recent action history. Instead, it is partly constructed through discussion, voting patterns, claims, accusations and the interpretations that other agents form from them~\cite{lan2024llm,meta2022human}. Social deduction games therefore provide a behavioural probe of language-mediated social reasoning, where competence is expressed not only in what an agent says, but in whether its messages and decisions change downstream beliefs, votes and outcomes~\cite{bailis2024werewolf,light2023avalonbench,lan2024llm}.

Most existing studies of LLM agents in such environments have treated the model, together with its prompt, memory or scaffold, as the object to be evaluated~\cite{light2023avalonbench,lan2024llm,bailis2024werewolf,chi2024amongagents, fan2024can,wang2023avalon}. This perspective has been useful for identifying behavioural signatures of pretrained agents: whether they can keep track of roles, infer hidden information, coordinate with partners, detect deception or produce persuasive messages~\cite{light2023avalonbench,lan2024llm,yoo2024finding,bailis2024werewolf, zhang2025dvm}. It also shows why social games are difficult benchmarks. A response can be locally fluent while failing to update beliefs, a vote can be individually reasonable while harming the team, and an apparently persuasive argument can have little effect on collective decisions. These observations suggest that social competence in LLM agents is better understood as a set of behaviourally grounded functions, rather than as a single property measured from isolated responses~\cite{akata2025playing,lan2024llm, willis2025will, xu2026csp4sdg}.

Reinforcement fine-tuning has become a central mechanism for turning pretrained language models into goal-directed agents~\cite{ma2024coevolving,cheng2024self,guo2025deepseek,zhai2025agentevolver,zhang2025evolvesearch, uehara2024understanding, wang2025parameter, zhang2025survey, zhang2025survey2, zhang2025survey3, zhang2025thinkfl, zhang2026towards}. Building on reinforcement learning from human feedback, preference optimization and related post-training methods, recent work has optimized LLMs with verifiable rewards, learned reward models, model-based evaluators and trajectory-level environment feedback~\cite{schulman2017proximal,ouyang2022training,rafailov2023direct,pan2026rlcsd,pan2025d}. These approaches have improved reasoning, coding, instruction following, tool use and long-horizon agent tasks, showing that post-training can reshape not only response style but also decision policies in interactive settings~\cite{shao2024deepseekmath, guo2025deepseek}.

RFT extends the study of LLMs in social deduction games from static evaluation to experience-driven learning~\cite{ma2024coevolving,cheng2024self,xu2024language}. In many RFT settings, feedback is relatively direct: a solution is correct or incorrect, code passes or fails tests, a tool-use trajectory succeeds or fails, or a response is preferred by a judge~\cite{ouyang2022training,rafailov2023direct,guo2025deepseek,xu2025learning}. In hidden-role games, by contrast, rewards are delayed, stochastic and entangled with other agents' beliefs and actions~\cite{lan2024llm,bailis2024werewolf,xu2024language}. An utterance may be fluent but fail to persuade; a vote may be locally reasonable but damage coordination; and a successful outcome may depend on how others interpret earlier messages. In this setting, RFT is not merely a post-training technique for improving task scores, but an operational model for studying how language agents acquire socially consequential behaviour through repeated interaction.

\section{Methods}\label{sec:method}

We describe the experimental and training framework used to study reinforcement fine-tuning of large language model agents in social deduction games. The methods are organized into three components. First, we introduce the AgentEvolver Game Arena, which generates interactive game trajectories, records hidden and public state, and converts natural-language interactions into structured training signals. Second, we describe the multi-agent social-cognitive reinforcement fine-tuning procedure used to optimize language-model agents for social reading and social influence objectives. Third, we outline a human-in-the-loop validation protocol that can test whether the model-only behavioural effects are recognizable and useful to human players.

\subsection{AgentEvolver Game Arena}

AgentEvolver Game Arena is a multi-agent environment for studying language-model behaviour in socially interactive games. The arena instantiates games with hidden roles, private observations, public discussion and consequential group decisions. It serves two purposes. First, it provides a controlled simulator in which LLM agents can interact with other model agents through natural language. Second, it converts these interactions into structured trajectories containing prompts, responses, parsed actions, hidden states, intermediate social outcomes and terminal game outcomes~\cite{zhang2024time,kang2022separation}.

The arena was designed to support both evaluation and reinforcement fine-tuning. During evaluation, it allows us to measure whether agents can infer hidden social information or influence collective decisions. During training, it provides trajectory-level rewards that can be assigned to specific players, roles or sides of the game. We describe the arena in three parts: the game environments, the multi-agent interaction protocol and the trajectory annotation procedure used to construct rewards and diagnostics.

\subsubsection{Social reasoning game environments}

We developed AgentEvolver Game Arena as a trajectory-generation environment for studying social cognition in language-model agents. The arena implements a set of partially observable multi-agent games in which agents must infer hidden information, communicate with other
players and make consequential decisions from dialogue. In this work, we used three families of social reasoning environments: One Night Werewolf, Classic Werewolf and Avalon. These games differ in horizon length, role structure and communication protocol, but share a common computational interface.

Each game instance is specified by a player set \(\mathcal{N}=\{1,\ldots,N\}\), a hidden role assignment \(z=(z_1,\ldots,z_N)\), a public game state \(s_t\), and a private observation \(o_{i,t}\) for each player \(i\) at time \(t\). At each decision point, a player receives a text prompt \(x_{i,t}\), generated from its private observation, the public history and the game rules, and
produces a natural-language response \(y_{i,t}\). The response is then parsed into a structured game action \(a_{i,t}\), such as a vote, a team proposal, an accusation or a role-specific night action. A complete episode therefore produces a trajectory

\begin{equation}
\tau =
\left\{
(z_i, x_{i,t}, y_{i,t}, a_{i,t}, s_t)
\right\}_{i\in\mathcal{N},\,t=1}^{T},
\end{equation}

together with terminal outcomes such as the winning side, the exiled player or the mission result.

The environments were designed to separate social-reading and social-influence demands. Social-reading tasks evaluate whether an agent can infer latent variables of the game, such as hidden roles or team membership, from the public dialogue and its private information. Social-influence tasks evaluate whether an agent can change subsequent group decisions, such as votes or team approvals, through its generated messages. This separation allows the same arena to support both diagnostic evaluation and reinforcement fine-tuning.

\subsubsection{Multi-agent interaction protocol}

Agents interact through a turn-based prompt-chaining protocol. At each turn, the arena constructs the prompt for the active player from a fixed game template, the player's role information and the public transcript available at that point in the game~\cite{zhang2025xraglog,zhang2024reducing}. The active model then samples a response according to

\begin{equation}
y_{i,t} \sim \pi_{\theta_i}(\cdot \mid x_{i,t}),
\end{equation}

where \(\pi_{\theta_i}\) denotes the policy assigned to player \(i\). The generated response is appended to the public transcript when the game phase is public, and is kept private when the action belongs to a hidden phase.

The arena supports both homogeneous and controlled-policy settings. In homogeneous evaluations, all players are assigned to the same model or API policy. In controlled-policy experiments, only a subset of players \(\mathcal{C}\subseteq\mathcal{N}\) is controlled by the trainable policy, while the remaining players are assigned to fixed opponent policies~\cite{zhang54optimization}:

\begin{equation}
\pi_i =
\begin{cases}
	\pi_{\theta}, & i \in \mathcal{C}, \\
	\pi^{\mathrm{opp}}_i, & i \notin \mathcal{C}.
\end{cases}
\end{equation}

This design makes it possible to evaluate whether training changes the behaviour of a target role, a target player or an entire side of the game, while holding the surrounding social environment fixed. For the full-game experiments, we use a same-side controlled-policy variant. Instead of updating only one focal player, the trainable policy is assigned to every role on a target side, such as all good-side players in Werewolf or all members of one side in Avalon. A single rollout therefore contains several controlled player trajectories from the same side. After the episode, each controlled player's prompts and responses are extracted as a role-conditioned training sample, while the side-level annotations are shared across those samples when appropriate. This reuse makes one interactive game provide multiple policy updates without mixing objectives across opposing sides.

For games with structured decisions, the model response is parsed into a machine-readable action. Invalid or missing actions are recorded as format failures and handled by the environment according to the predefined evaluation protocol. We log both the raw language output and the parsed action, allowing downstream analyses to distinguish failures of reasoning, persuasion and action formatting~\cite{zhang2025scalalog,zhang2026efficient,he2025united}.

\subsubsection{Trajectory and reward annotation}

After each episode, the arena annotates the trajectory with task-specific and game-level outcomes. Let \(m_k(\tau)\) denote an annotated metric computed from trajectory \(\tau\), such as hidden-role prediction accuracy, target vote hit rate, team approval rate or side win rate. The arena distinguishes optimization rewards from diagnostic metrics. For outcome-only baseline experiments, the terminal side outcome can be used directly as the scalar reward. For the online same-side social-cognitive experiments, however, terminal win rate is logged but not used as the primary training reward, because it produced a sparse and unstable learning signal in preliminary online runs. Training rewards are therefore defined as weighted combinations of task-aligned social-cognitive metrics,

\begin{equation}
R(\tau) = \sum_{k} \lambda_k m_k(\tau),
\end{equation}

where the weights \(\lambda_k\) specify the behavioural objective of the experiment. For example, social-reading rewards emphasize correct inference of hidden roles, whereas social-influence rewards emphasize whether subsequent collective decisions move in the direction intended by the controlled agent or side.

In addition to the scalar reward used for optimization, the arena stores a richer set of diagnostic annotations for analysis~\cite{zhang2024multivariate,zhang2025logdb,zhang2025log}. These include format-validity rates, response length, vote distributions, protection or targeting rates for specific roles, and terminal win rates. Terminal win rate is therefore treated primarily as an external behavioural outcome: it tests whether improvements in the optimized social-cognitive objectives transfer to actual game success, rather than defining the online objective itself.

This trajectory-level interface provides a common basis for reinforcement fine-tuning and evaluation across games. The same game engine generates language prompts, executes parsed actions, records hidden and public state, computes rewards and exports full trajectories for analysis~\cite{zhang2024towards,zhang2026runtimeslicer,liu2025aaad}. Thus, AgentEvolver Game Arena links natural-language social interaction with structured reinforcement-learning signals while preserving the multi-agent dynamics of the underlying games.

\subsection{Multi-Agent Social-Cognitive Reinforcement Fine-Tuning}

We fine-tune language-model agents with reinforcement signals that target two complementary components of social cognition: \emph{social reading}, the ability to infer latent social variables from dialogue and partial observations, and \emph{social influence}, the ability to alter other players' subsequent beliefs or decisions through communication. These two abilities are closely related but not identical. A player may correctly infer hidden roles without successfully persuading others, or may temporarily influence a vote without maintaining an accurate belief state. We therefore model them as separate reward components and later combine them into a same-side social-cognitive objective whose effect on terminal game outcomes is evaluated separately.

For each episode \(\tau\), the arena records prompts, responses, parsed actions, private role information, public transcripts, intermediate social outcomes and terminal game outcomes. A controlled player \(i\in\mathcal{C}\) generates responses according to
\begin{equation}
	\label{eq:sc-rft-policy}
	y_{i,t} \sim \pi_{\theta}(\cdot \mid x_{i,t}) ,
\end{equation}
where \(x_{i,t}\) is the prompt at time \(t\), \(y_{i,t}\) is the generated language response and \(\pi_{\theta}\) is the trainable policy. The response is parsed into a structured action when the game phase requires one, while the raw language is preserved for behavioural analysis. Non-controlled players are assigned fixed opponent policies. In same-side training, \(\mathcal{C}\) contains all trainable players on the selected side. The same policy parameters are shared across these players, but each update is conditioned on the player's private role, public history and phase-specific prompt. Thus, a rollout can yield multiple controlled-player training sequences while preserving role-specific information.

\subsubsection{Social reading objective}

The social reading objective rewards the agent for inferring hidden social state from the observed interaction history. In social deduction games, the relevant hidden variables include team membership, private role information, night actions and latent strategic intentions. These variables are not directly available to the agent, but they leave traces in public discussion, voting patterns and the sequence of deaths or mission outcomes. The reading objective is intended to measure whether the agent can recover this latent structure from the information available to its role.

At selected decision points, the agent emits a structured belief state \(b_{i,t}\) in addition to its public response or action. Depending on the game and phase, this belief state may include suspected adversaries, trusted players, role hypotheses or predicted future actions. The simulator compares the belief state with the hidden state available to the arena but not to the agent. For each annotated reading event, the comparison is performed by a task-specific scoring function \(m_{\mathrm{read}}\). For set-valued predictions, such as the set of suspected werewolves, \(m_{\mathrm{read}}\) is the F1 score between the predicted and true target sets; for categorical role predictions, it is exact-match accuracy.

The episode-level reading reward averages these scores over all annotated reading events:
\begin{equation}
	\label{eq:social-reading-reward}
	R_{\mathrm{read}}(\tau)
	=
	\frac{1}{|\mathcal{E}_{\mathrm{read}}|}
	\sum_{(i,t)\in\mathcal{E}_{\mathrm{read}}}
	m_{\mathrm{read}}
	\left(
	b_{i,t}, z^{\mathrm{target}}_t
	\right),
\end{equation}
where \(\mathcal{E}_{\mathrm{read}}\) is the set of reading decisions in trajectory \(\tau\), and \(z^{\mathrm{target}}_t\) denotes the hidden target variable queried at time \(t\). This objective encourages the policy to maintain explicit beliefs about latent social structure rather than only producing fluent or plausible dialogue. It also provides an intermediate learning signal before the final game outcome is known.

\subsubsection{Social influence objective}

The social influence objective rewards the agent when its communication moves subsequent group behaviour in an intended direction. In contrast to social reading, which evaluates whether the agent correctly represents hidden state, social influence evaluates whether the agent's message changes the behaviour of other players. This is central to social deduction games: private beliefs are only useful if they can be converted into public coordination, credible persuasion or strategic deception.

For each controlled public message, the arena records the agent's intended social target, denoted \(u_{i,t}\). This target may correspond to pressuring a suspected adversary, encouraging approval of a team, redirecting suspicion away from an ally or coordinating a vote. The arena then records later actions by other players, denoted \(A_{-i,t+1}\). These actions include votes, approvals, accusations, team selections or other game-specific decisions. The episode-level influence reward is
\begin{equation}
	\label{eq:social-influence-reward}
	R_{\mathrm{inf}}(\tau)
	=
	\frac{1}{|\mathcal{E}_{\mathrm{inf}}|}
	\sum_{(i,t)\in\mathcal{E}_{\mathrm{inf}}}
	m_{\mathrm{inf}}
	\left(
	u_{i,t}, A_{-i,t+1}, z
	\right),
\end{equation}
where \(m_{\mathrm{inf}}\) is a task-specific metric measuring whether subsequent decisions moved toward the intended social outcome.

The precise form of \(m_{\mathrm{inf}}\) depends on the game and side. In Werewolf, good-side influence is measured by whether other players follow the controlled agent toward exiling a wolf, whereas wolf-side influence is measured by whether good players misdirect suspicion toward other good players or avoid voting for wolves. In Avalon, analogous influence signals can be defined over team proposals, team approvals and mission-related suspicion. Thus, the influence objective measures a causal social function of language: whether the message changes later strategic behaviour, not merely whether the message is coherent.

\subsubsection{Joint multi-agent social-cognitive reward}

The reading and influence objectives capture different mechanisms of social competence. Reading rewards the agent for identifying latent social structure; influence rewards the agent for changing public behaviour. Full-game success, however, also depends on using these abilities in a coordinated way across the side. We therefore combine the two social-cognitive objectives with auxiliary game-specific intermediate events, while treating terminal game outcome as a diagnostic measure rather than as the default online reward.

Let \(W(\tau)\in\{0,1\}\) indicate whether the controlled side wins the game. We log \(W(\tau)\) for evaluation. The joint same-side social-cognitive reward used for optimization is
\begin{equation}
	\label{eq:joint-social-cognitive-reward}
	\begin{aligned}
		R(\tau)
		=&\ \lambda_{\mathrm{read}} R_{\mathrm{read}}(\tau)
		+ \lambda_{\mathrm{inf}} R_{\mathrm{inf}}(\tau)
		+ \lambda_{\mathrm{aux}} R_{\mathrm{aux}}(\tau) \\
		&- \lambda_{\mathrm{fmt}} P_{\mathrm{fmt}}(\tau)
		- \lambda_{\mathrm{leak}} P_{\mathrm{leak}}(\tau).
	\end{aligned}
\end{equation}
Here \(R_{\mathrm{aux}}\) contains game-specific intermediate outcomes, such as exiling an adversary, protecting a teammate, selecting a favourable team or causing a desired collective decision. \(P_{\mathrm{fmt}}\) penalizes invalid structured outputs, and \(P_{\mathrm{leak}}\) penalizes illegal public disclosure of hidden information, such as revealing private team knowledge in a public phase.

This reward decomposition serves two purposes. First, it avoids the sparsity of terminal game outcomes by providing intermediate social-cognitive feedback. Second, it enables diagnostic analysis. A model may improve its win rate because it reads hidden roles more accurately, because it better persuades other players, because it produces more valid actions, or because it exploits a formatting shortcut. By logging all reward components and terminal outcomes separately, we can distinguish these possibilities and test whether apparent performance gains correspond to meaningful changes in social behaviour.

\subsubsection{GRPO optimization}

We optimize the trainable policy using Group Relative Policy Optimization. For each task instance, the current policy samples a group of \(G\) trajectories and receives a scalar reward for each trajectory. These trajectories correspond to different stochastic continuations of the same or comparable game condition, allowing the reward of one trajectory to be interpreted relative to other sampled behaviours from the same policy. GRPO estimates a normalized relative advantage within the group:
\begin{equation}
	\label{eq:grpo-advantage}
	\hat{A}_g
	=
	\frac{
		R(\tau_g)-\mu_R
	}{
		\sigma_R+\epsilon
	},
	\qquad
	\mu_R
	=
	\frac{1}{G}
	\sum_{g=1}^{G}
	R(\tau_g),
\end{equation}
where \(\mu_R\) and \(\sigma_R\) are the mean and standard deviation of rewards in the group.

The policy is updated with a clipped policy-gradient objective regularized by a KL penalty to the reference model:
\begin{equation}
	\label{eq:grpo-objective}
	\begin{aligned}
		\mathcal{L}_{\mathrm{GRPO}}(\theta)
		=
		-\frac{1}{G}
		\sum_{g=1}^{G}
		\sum_{t}
		\min
		\Big[
		&\rho_{g,t}(\theta)\hat{A}_g,\,
		\mathrm{clip}\big(\rho_{g,t}(\theta),1-\epsilon,1+\epsilon\big)\hat{A}_g
		\Big] \\
		&+
		\beta\,\mathrm{KL}
		\left(
		\pi_{\theta}\,\|\,\pi_{\mathrm{ref}}
		\right),
	\end{aligned}
\end{equation}
where the likelihood ratio is
\begin{equation}
	\label{eq:grpo-ratio}
	\rho_{g,t}(\theta)
	=
	\frac{
		\pi_{\theta}(y_{g,t}\mid x_{g,t})
	}{
		\pi_{\theta_{\mathrm{old}}}(y_{g,t}\mid x_{g,t})
	}.
\end{equation}
The KL term limits drift from the reference model and helps preserve language quality, instruction following and action formatting. GRPO is particularly suitable for our setting because social-deduction rewards are sparse, noisy and trajectory-dependent, and because a learned value function would need to model complex multi-agent dynamics. Group-relative normalization provides a practical variance-reduction mechanism without introducing a separate critic.

\subsubsection{Training regimes}

We study three training regimes. In the \emph{social reading} regime, the reward emphasizes hidden-state inference. This setting tests whether an agent can learn to infer latent roles or team membership from partial observations and public dialogue. In the \emph{social influence} regime, the reward emphasizes downstream behavioural change. This setting tests whether the agent can learn messages that alter votes, approvals or suspicion in a desired direction. In the \emph{joint social-cognitive} regime, both social-cognitive components are combined with auxiliary intermediate game events:
\begin{equation}
	\label{eq:joint-regime}
	\begin{aligned}
		R_{\mathrm{joint}}(\tau)
		=&\ \lambda_{\mathrm{read}} R_{\mathrm{read}}(\tau)
		+ \lambda_{\mathrm{inf}} R_{\mathrm{inf}}(\tau)
		+ \lambda_{\mathrm{aux}} R_{\mathrm{aux}}(\tau) \\
		&- \lambda_{\mathrm{fmt}} P_{\mathrm{fmt}}(\tau)
		- \lambda_{\mathrm{leak}} P_{\mathrm{leak}}(\tau).
	\end{aligned}
\end{equation}

During training, the trainable policy controls a specified player, role or side, while the remaining players are fixed opponent policies. In same-side runs, all controlled roles on the selected side are optimized together from the same rollout, and each controlled role contributes its own prompt-response sequence to the GRPO batch. This makes it possible to ask whether a particular social-cognitive capability improves when one side of the social system is optimized as a coordinated policy. During evaluation, we run independent held-out games rather than reusing the training reward as the sole metric. We report side win rate, role-specific win rate, valid-action rate, reading accuracy, vote distributions and influence-following rates. This separation between training rewards and evaluation metrics allows us to test whether reinforcement fine-tuning produces robust social behaviour rather than merely optimizing a particular reward function~\cite{huang2025uda,duan2025logaction,xiao2025coorlog}.

\subsection{Human evaluation}

We conducted a blinded evaluation of complete nine-player Werewolf
trajectories to determine whether the behavioural changes produced by
social-cognitive reinforcement fine-tuning were recognizable at the level of
complete games. The evaluation compared trajectories generated by the base
policy and the reinforcement-fine-tuned policy. In both conditions, the
candidate policy controlled all good-side players, while the wolf side was
controlled by fixed opponent models.

The human-evaluation component of this study was conducted in accordance
with the relevant guidelines and regulations. The study was reviewed by the
Ethics Committee and was determined to be exempt from
ethics review. Informed consent was obtained from all human participants.

\subsubsection{Trajectory selection}

We constructed matched base--RFT trajectory pairs under the same game
identifier and opponent condition. Candidate trajectories were retained only
when they contained complete model-call histories and public event records,
contained no fallback public speech, and satisfied the predefined
structured-output validity criterion. This filtering procedure was applied
identically to both policy conditions.

\subsubsection{Blinded evaluation procedure}

Evaluators reviewed complete game trajectories containing the role
configuration, public discussion, votes, eliminations, night actions, and the
private belief and strategic-intent records of the candidate-controlled
good-side players. The identity and training condition of the candidate policy
were hidden during evaluation.

Each trajectory was rated on a five-point integer scale along five dimensions:
overall value, social-reading quality, social-influence and coordination,
human-like gameplay, and strategic correctness. Evaluators were instructed to
assess whether the agents' beliefs were supported by the evidence available at
the time, whether public communication and votes were consistent with their
private reasoning, and whether the resulting collective decisions were
strategically appropriate. They were explicitly instructed not to assign a
high score solely because the good side won.

All ratings were analysed at the trajectory-rating level. The same evaluation
rubric, scoring interface, and trajectory presentation format were used across
policy conditions.

\subsubsection{Statistical analysis}

For each evaluation dimension, we calculated descriptive statistics separately
for the base and reinforcement-fine-tuned conditions. Uncertainty intervals
were estimated by non-parametric bootstrap resampling over trajectory ratings.

We additionally performed matched comparisons between base and RFT
trajectories generated under the same game identifier and opponent condition.
For each evaluation dimension, we recorded whether the RFT trajectory received
a higher, equal, or lower score than its matched base trajectory. Statistical
significance was assessed using two-sided paired sign-flip permutation tests.

As a robustness check, the matched analysis was repeated after excluding
trajectories involving an opponent model that overlapped with the evaluator
model family.

\section*{Data availability}

All participant and model simulation data from the experiments are available via GitHub at \url{https://anonymous.4open.science/r/RL-Game-Data-Availability}.

\section*{Code availability}

The code underlying this study, prompt variations and model simulations is available via GitHub at \url{https://anonymous.4open.science/r/RL-Game}.

\backmatter

\bibliography{sn-bibliography}

\end{document}